\documentclass[journal]{IEEEtran}
\usepackage{graphicx}
\usepackage{times}
\usepackage{booktabs}
\usepackage{latexsym}
\usepackage{booktabs} 
\usepackage{bm}
\usepackage{hyperref}
\usepackage{xspace}
\usepackage{amsmath}
\usepackage{makecell}
\usepackage{multirow}
\usepackage{amsfonts}
\usepackage{algorithmic}
\usepackage{algorithm}
\usepackage{subfigure}
\usepackage{enumitem}
\usepackage{url}
\usepackage[flushleft]{threeparttable}
\usepackage{longtable}
\usepackage{supertabular}
\usepackage{bm}
\usepackage{xcolor}
\usepackage{multirow}
\usepackage{tabularx}
\usepackage{cite}
\usepackage{graphicx}
\usepackage[normalem]{ulem}
\useunder{\uline}{\ul}{}

\begin{document}
\title{Explainable Artificial Intelligence Methods in Combating Pandemics: A Systematic Review}
\author{Felipe~Giuste$^*$,~\IEEEmembership{Student Member,~IEEE},
        Wenqi~Shi$^*$,~\IEEEmembership{Student Member,~IEEE},
        Yuanda~Zhu,~\IEEEmembership{Student Member,~IEEE},
        Tarun~Naren,
        Monica~Isgut,
        Ying~Sha,
        Li~Tong,
        Mitali~Gupte,
        and~May~D.~Wang,~\IEEEmembership{Fellow,~IEEE}% <-this % stops a space
\thanks{* The ﬁrst two authors contributed equally to this work.}
\thanks{F. Giuste, Y. Sha, L. Tong, M. Gupte, and M. D. Wang are with the Wallace H. Coulter School of Biomedical Engineering, Georgia Institute of Technology, Emory University, Atlanta, GA 30322 USA (e-mail: fgiuste@gatech.edu; ysha8@gatech.edu; ltong@gatech.edu; mitali.gupte@gatech.edu; maywang@gatech.edu).}% <-this % stops a space
\thanks{W. Shi and Y. Zhu are with the Department of Electrical and Computer Engineering, Georgia Institute of Technology, Atlanta, GA, 30332 USA (e-mail: wshi83@gatech.edu; yzhu94@gatech.edu).}% <-this 
\thanks{M. Isgut is with the School of Biology, Georgia Institute of Technology, Atlanta, GA 30322 USA (email: misgut@gatech.edu).}
\thanks{T. Naren is with the Department of Nuclear and Radiological Engineering, Georgia Institute of Technology, Atlanta, GA, 30332 USA (e-mail: tnaren3@gatech.edu).}% <-this % stops a space
\thanks{Corresponding author: May D. Wang (email: maywang@gatech.edu).}% <-this % stops a space
%\thanks{Manuscript received April 19, 2005; revised December 27, 2012.}
}

% The paper headers
\markboth{IEEE Journal,~Vol.~XX, No.~X, December~2021}%
{Giuste \MakeLowercase{\textit{et al.}}: Explainable Artificial Intelligence Methods in Combating Pandemics: A Systematic Review}

\maketitle

% abstract

\begin{abstract}
Despite the myriad peer-reviewed papers demonstrating novel Artificial Intelligence (AI)-based solutions to COVID-19 challenges during the pandemic, few have made significant clinical impact. The impact of artificial intelligence during the COVID-19 pandemic was greatly limited by lack of model transparency. This systematic review examines the use of Explainable Artificial Intelligence (XAI) during the pandemic and how its use could overcome barriers to real-world success. We find that successful use of XAI can improve model performance, instill trust in the end-user, and provide the value needed to affect user decision-making. We introduce the reader to common XAI techniques, their utility, and specific examples of their application. Evaluation of XAI results is also discussed as an important step to maximize the value of AI-based clinical decision support systems. We illustrate the classical, modern, and potential future trends of XAI to elucidate the evolution of novel XAI techniques. Finally, we provide a checklist of suggestions during the experimental design process supported by recent publications. Common challenges during the implementation of AI solutions are also addressed with specific examples of potential solutions. We hope this review may serve as a guide to improve the clinical impact of future AI-based solutions. 
\end{abstract}

\begin{IEEEkeywords}
Explainable Artificial Intelligence, COVID-19, Explanation Generation, Explanation Representation, Explanation Evaluation, Medical Imaging, Electronic Health Records
\end{IEEEkeywords}

\IEEEpeerreviewmaketitle

\section{Introduction}
\begin{figure*}[]
    \centering
    \includegraphics[width=\linewidth]{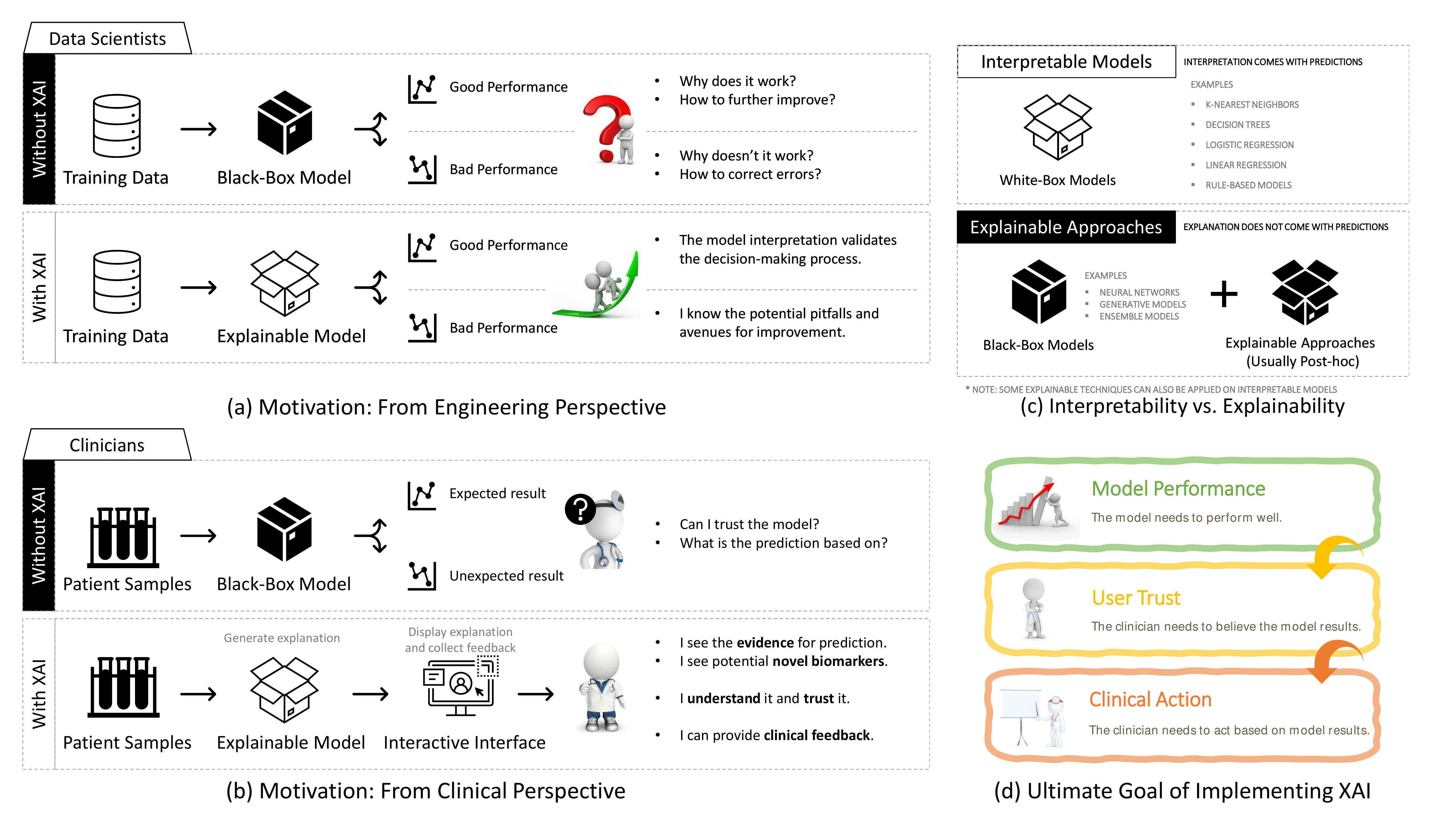}
    \vspace{-10mm}
    \caption{Problem statement and motivation of XAI in clinical applications. (a) From a model development perspective, XAI techniques enhance the transparency of AI models, allowing for more confident clinical decision-making and increasing the real-world utility of AI approaches. (b) From a clinical perspective, clinicians can benefit from XAI by gaining insight into how the AI models reach solutions from clinical data. (c) The term "interpretability" refers to a property of AI systems in which the process by which they arrive at a conclusion is easily understood. K-nearest neighbors, decision trees, logistic regression, linear models, and rule-based models are all popular interpretable machine learning methods. Explainable AI is frequently used to refer to methods (usually post-hoc) for enhancing comprehension of black-box models such as neutral networks and ensemble models. Explainable AI methods attempt to summarize the rationale for a model's behavior or to generate insights into the underlying cause of a model's decision. Both interpretability and explainability are frequently used interchangeably, and both seek to shed light on the model's credibility. In this review, we will focus on XAI methods used in clinical settings. (d) AI-based clinical solutions should meet three criteria: achieve high performance, instill user trust, and generate user response, all of which demonstrate the importance of XAI in clinical applications.}
    \label{fig:overview}
\end{figure*}

\IEEEPARstart{C}{oronavirus} disease 2019 (COVID-19) has become a worldwide phenomenon with over 272 million cases and claiming over four million lives \cite{hu2020characteristics}. Medical imaging and clinical data have been explored as potential supplements to molecular test screening of patients potentially infected with COVID-19 \cite{rubin2020role, soltan2020rapid, yan2020interpretable}. The need for fast COVID-19 detection has led to a massive number of Artificial Intelligence (AI) solutions to alleviate this clinical burden. Unfortunately, very few have succeeded in making a real impact \cite{Heaven2021-ef}. As the world slowly transitions from disease detection and containment to maximizing patient outcomes, so too must AI solutions meet this urgent requirement. To face these new challenges, we must look back to identify areas of improvement in AI. For example, how many of the wildly successful models published in the last year actually made a meaningful clinical impact? A major barrier limiting the real-world utility of AI approaches in clinical decision support is the lack of interpretability in their results.

A wide range of imaging and non-imaging clinical data sources are used to extract meaningful patterns for automated COVID detection tasks. The most common data sources used for elucidating COVID-19 pathology include X-ray, computed tomography (CT) and electronic health records. 

Using imaging and clinical features (including past medical history, current medications, and demographic data), state of art AI models have achieved impressive performance \cite{soltan2020rapid, zhang2020clinically, li2020using, minaee2020deep}. This is due to the complex data transformations applied to the input data and the large numbers of parameters optimized by the model without human intervention. This approach can lead to unexpected or inaccurate results \cite{Rahman2021-ax}. This is especially true if the model learns features unrelated to the challenge being solved, as can be the case when significant confounders, such as in patient demographic or underlying disease epidemiology effects, are present within training datasets. Thus, physicians and other healthcare practitioners are often reluctant to trust many high-performing yet unintelligible AI systems. Additionally, lack of interpretability in AI algorithms limits the ability of researchers and model developers to identify potential pitfalls and avenues for improvement. State-of-the-art approaches may be finding non-optimal solutions, or worse, basing their solutions on irrelevant input features \cite{su2019one, eykholt2018robust}.

Explainable artificial intelligence (XAI) is a collection of processes and methods that enables human users to comprehend and trust machine learning algorithms' results \cite{alvarez2018towards}. XAI techniques improve the transparency of AI models, thus facilitating confident clinical decision-making and increasing the real-world utility of AI approaches. Clinicians benefit from XAI by gaining insight into how the AI models reach solutions from clinical data, as shown in Fig. \ref{fig:overview}. AI-based clinical solutions should meet three criteria: achieve high performance, instill user trust, and generate user response. Specifically, the model should have achieved sufficient performance at their task on a real-world dataset not used during the training process in order to be considered for real-world use. Guidelines for establishing and reporting real-world clinical trial performance could be found in the SPIRIT-AI \cite{Cruz_Rivera2020-al} and CONSORT-AI \cite{Liu2020-jm} guidelines. Trust in the AI-solution may be established with XAI, especially when visual feedback is provided to the user on important metrics used to obtain the model prediction. Finally, no solution is effective if it does not result in a change in user response. This response may include a change in treatment plan, patient prioritization, or diagnosis. This response must be consistent with clinical expertise and evidence-based protocols. This review primarily focuses on the XAI solutions affecting user trust, but model performance and user interfaces are also mentioned where appropriate. XAI can allow for validation of extracted features, confirm heuristics, identify subgroups, and generate novel biomarkers \cite{makino2020differences}. In addtion, XAI can also support research conclusions and guide research field advancement by identifying avenues of model performance improvement. 

\begin{figure*}
    \centering
    \includegraphics[width=\linewidth]{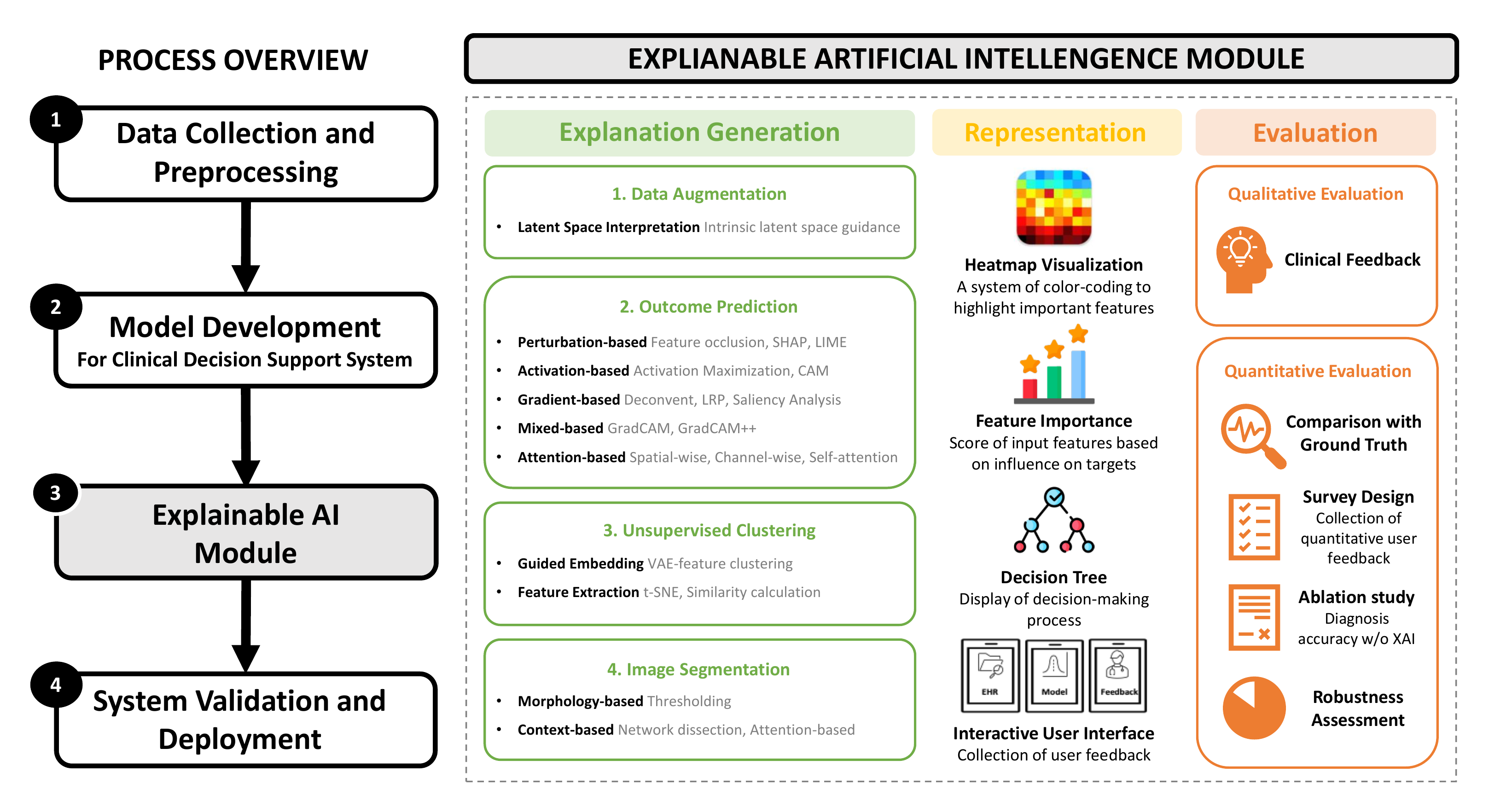}
    % \vspace{-5mm}
    \caption{Overview of paper structure. XAI techniques increase the transparency of AI models, enabling more confident clinical decision-making and increasing the practical utility of AI approaches. In this paper, we present a comprehensive review of explanation generation, representation, and evaluation methods used to support AI-enabled clinical decision support systems during the COVID-19 pandemic.}
    \label{fig:methods}
\end{figure*}

In this systematic review, we describe the current usage of XAI techniques to solve COVID-19 clinical challenges. Upon review of the current literature leveraging AI for COVID-19 detection and risk assessment, we provide strong support for their increased use if clinical integration is the goal. The remaining of this paper is structured as follows (see Fig. \ref{fig:methods}): Section \ref{sec:PRISMA} illustrates how the XAI-based studies applicable to COVID-19 were selected using the Preferred Reporting Items on Systematic Reviews and Meta-analysis (PRISMA) model and exclusivity criteria; Section \ref{sec:method} provides a comprehensive overview of the XAI approaches used to support AI-enabled clinical decision support systems during COVID-19 pandemic; Section \ref{sec:representation} and Section \ref{sec:evaluation} describes the representation of explanations and evaluation of XAI methods; Section \ref{sec:discussion} and \ref{sec:conclusion} summarize the contribution of this paper, provide a schema of the integration of explainable AI module in both model development and clinical practice, and discuss potential challenges and future work of XAI.

\section{Paper Selection}
\label{sec:PRISMA}

\begin{figure}[]
    \centering
    \includegraphics[width=\linewidth]{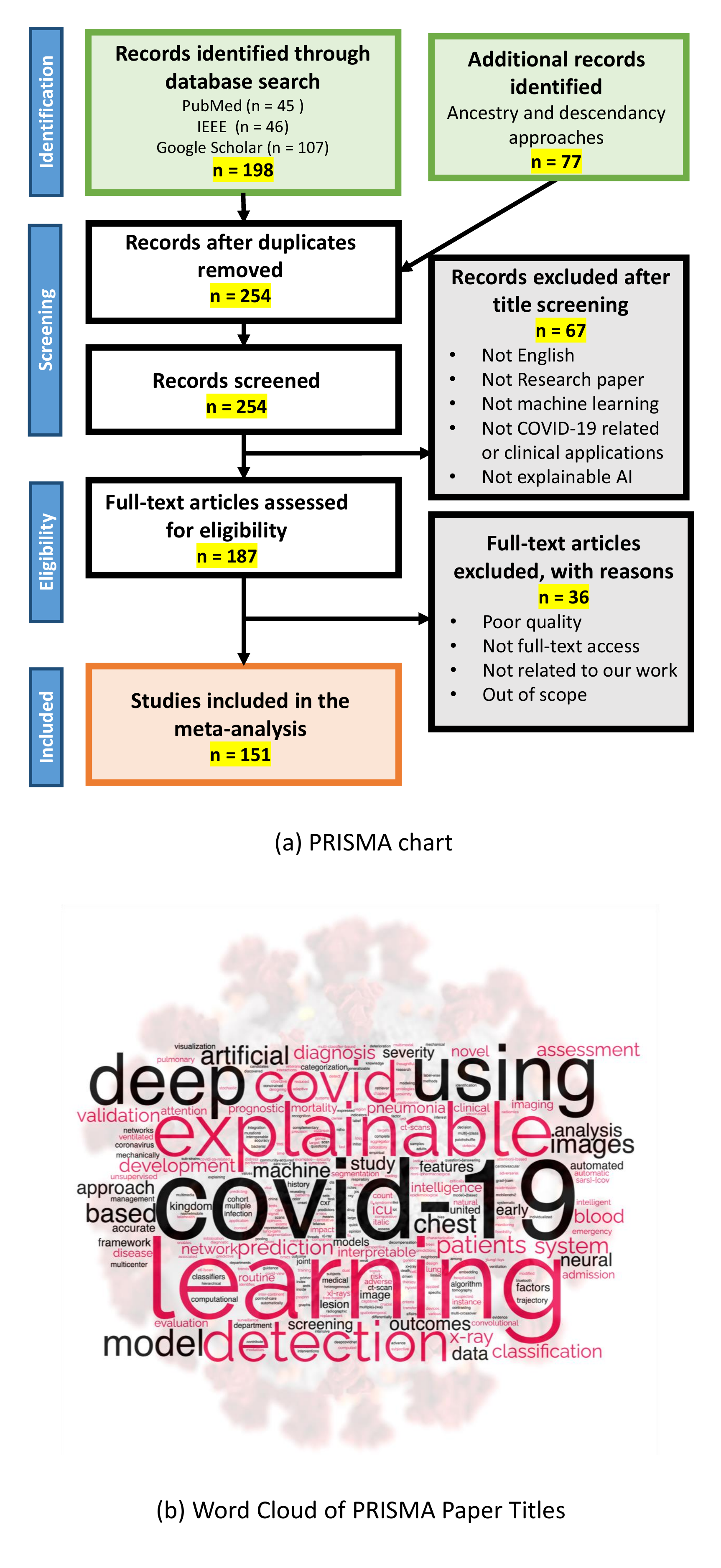}
    \vspace{-10mm}
    \caption{PRISMA chart for systematic paper selection and quality assessment.}
    \label{fig:prisma}
\end{figure}

This systematic review paper follows the PRISMA guidelines \cite{moher2009preferred}, as shown in Fig. \ref{fig:prisma} (a). PRISMA is an evidence-based minimum set of items for systematic reviews and meta-analyses with the goal of assisting authors in improving their literature review. 

We first conducted a systematic search of papers using three large public databases, PubMed, IEEE Xplore and Google Scholar. The keywords for our paper identifications are: \textit{(deep learning OR machine learning OR artificial intelligence) AND (interpretable OR explainable OR interpretable artificial intelligence OR explainable artificial intelligence) AND (COVID-19 OR SARS-CoV-2)}. We limited the date of publication between January 2020 and October 2021 to reflect the most recent progress. We selected this search approach with the help of Emory University librarians to ensure adequate breadth and specificity of search results. The word cloud generated by titles of selected article can also indicate our focus on XAI and COVID-19, as shown in Fig. \ref{fig:prisma} (b).

Our initial search matched 45 papers in PubMed, 46 in IEEE Xplore, 107 in Google Scholar, and 77 from external resources. After identifying initial papers, we then conducted a manual screening by reading the titles and abstracts and eliminating 67 papers that did not discuss XAI or COVID-19. Among the 187 remaining papers, we further excluded 36 papers for other reasons such as quality. In total, we included 151 papers from PRISMA in this study. These studies were thoroughly analyzed in order to gain a better understanding of XAI approaches, which is critical for model interpretation and has become a necessary component for any AI-based approach seeking to make a clinical impact.

\begin{figure*}[]
    \centering
    \includegraphics[width=\linewidth]{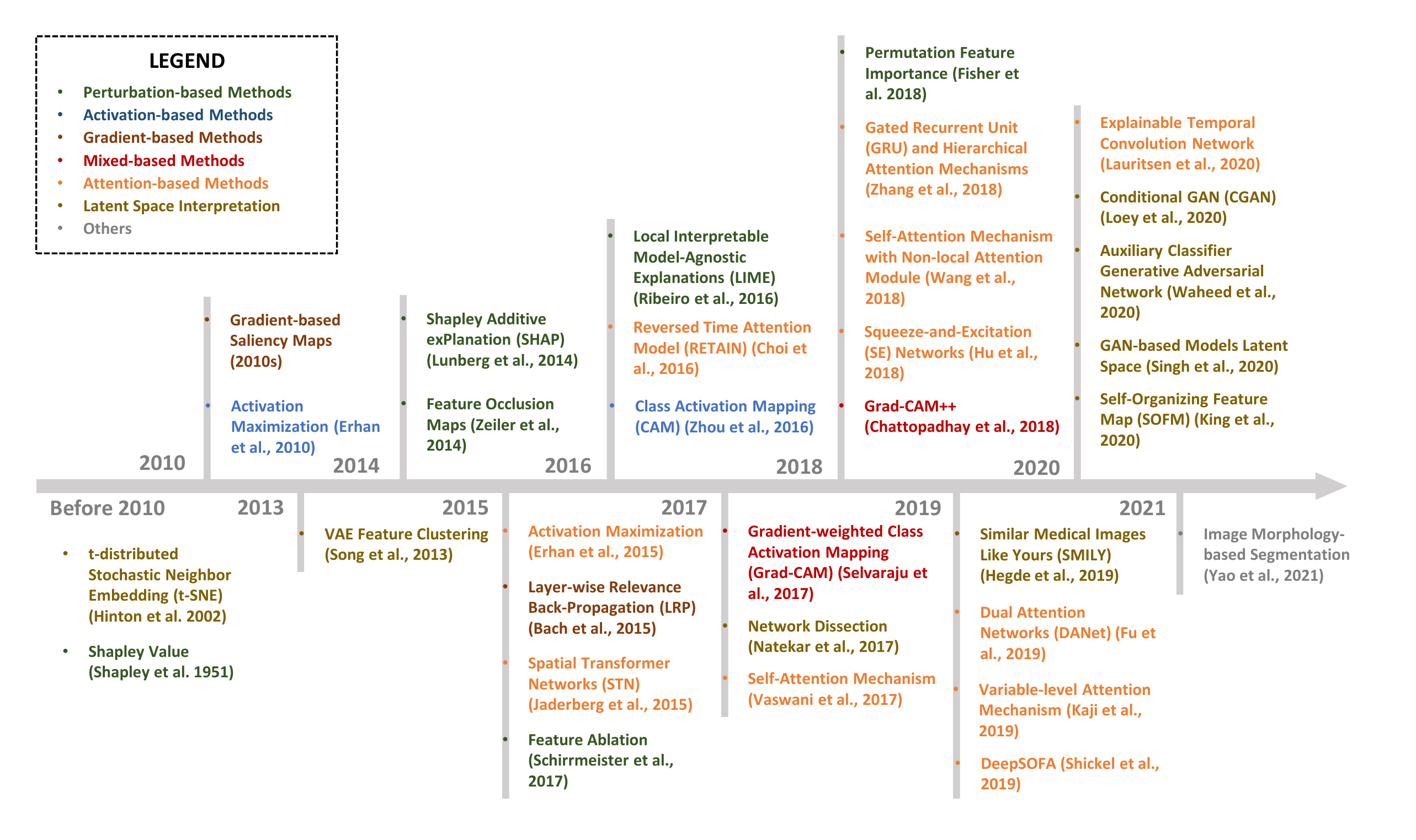}
    \vspace{-8mm}
    \caption{A brief summary of significant milestones in the development of XAI methods. According to their underlying theory, we classified these popular XAI methods into six categories: perturbation-based, activation-based, gradient-based, mixed-based, attention-based, and latent space interpretation. In the early stages of XAI development, perturbation-based, activation-based, and gradient-based methods are critical for model interpretation and generation of explanations. Recent years have seen significant advancements in mixed-based methods (combination of activation- and gradient-based methods), attention mechanisms, and latent space interpretations, all of which have played a significant role in medical XAI.}
    \label{fig:timeline}
\end{figure*}

\section{XAI Methods in COVID-19 Applications}
\label{sec:method}

In this section, we introduce XAI approaches used to support AI-enabled clinical decision support systems during the COVID-19 pandemic. We categorize them as follows: data augmentation, outcome prediction, unsupervised clustering, and image segmentation. Moreover, we organized XAI methods according to the underlying theory within each task, as shown in Fig. \ref{fig:timeline}. Additional technical details and clinical applications will be discussed below. An overview of XAI methods implemented in clinical applications is presented in Fig. \ref{fig:methods}. In addition, we also summarized publicly available COVID-19-related imaging (Table \ref{tab:imaging}) and non-imaging datasets (Table \ref{tab:non-imaging}) that were used in these clinical applications.

\begin{table*}[]
\centering
\caption{Publicly Available COVID-19 Medical Radiology Imaging Datasets}
\label{tab:imaging}
\resizebox{\textwidth}{!}{%
\begin{tabular}{@{}l|l|l|l|l|l@{}}
\toprule
\multicolumn{1}{c|}{\textbf{Dataset}} & \multicolumn{1}{c|}{\textbf{Size}} & \multicolumn{1}{c|}{\textbf{Format}} & \multicolumn{1}{c|}{\textbf{Modality}} & \multicolumn{1}{c|}{\textbf{Country}} & \multicolumn{1}{c}{\textbf{Reference}} \\ \midrule
Coronacases.ORG & 10 patients & Online & CT & China & {\ul https://coronacases.org/} \\ \midrule
COVID-CT (UCSD) & 349 images from 216 patients & PNG & CT & Global & {\ul https://github.com/UCSD-AI4H/COVID-CT} \\ \midrule
COVID-19 CT Lung and Infection Segmentation Dataset & 20 cases & NIfTI & CT & Global & {\ul https://zenodo.org/record/3757476} \\ \midrule
COVID-19 Segmentation Dataset & 100 images from 40 patients & NIfTI & CT & Italy & {\ul http://medicalsegmentation.com/covid19} \\ \midrule
MosMed COVID-19 Chest CT database & 1110 patients & NIfTI & CT & Russia & {\ul https://mosmed.ai/datasets/covid19\_1110} \\ \midrule
Società Italiana di Radiologia Medica (SIRM) & 68 patients & JPG & X-ray & Italy & {\ul https://sirm.org/covid-19/} \\ \midrule
AIforCOVID & 983 patients & DICOM & X-ray & Italy & {\ul https://aiforcovid.radiomica.it/} \\ \midrule
British Society or Thoracic Imaging (BSTI) & 59 patients & Online & X-ray & UK & {\ul https://bit.ly/BSTICovid19\_Teaching\_Library} \\ \midrule
COVID-19 Image Collection & 930 images from 461 patients & JPG, PNG & X-ray, CT & Global & {\ul https://github.com/ieee8023/covid-chestxray-dataset} \\ \midrule
BIMCV-COVID19+ & 7377 X-ray and 6687 CT studies & DICOM & X-ray, CT & Global & {\ul https://bimcv.cipf.es/bimcv-projects/bimcv-covid19/} \\ \midrule
Radiopaedia COVID-19 & 101 patients & JPG & X-ray, CT & Global & {\ul https://radiopaedia.org/articles/covid-19-3} \\ \midrule
Eurorad & 50 patients & JPG & X-ray, CT & Global & {\ul https://www.eurorad.org/} \\ \bottomrule
\end{tabular}%
}
\end{table*}

\begin{table*}[]
\centering
\caption{Publicly Available COVID-19 Non-imaging Data Platform}
\label{tab:non-imaging}
\resizebox{\textwidth}{!}{%
\begin{tabular}{@{}l|l|l|l|l@{}}
\toprule
\multicolumn{1}{c|}{\textbf{Dataset}} & \multicolumn{1}{c|}{\textbf{Source}} & \multicolumn{1}{c|}{\textbf{Format}} & \multicolumn{1}{c|}{\textbf{Country}} & \multicolumn{1}{c}{\textbf{Reference}} \\ \midrule
UCSF COVID-19 Research Data Mart & UCSF & \begin{tabular}[c]{@{}l@{}}1. Structured EHR data\\ 2. Clinical notes\\ 3. Imaging data\end{tabular} & United States & {\ul https://data.ucsf.edu/covid19} \\ \midrule
All of Us: COVID-19 research initiative & All of US & \begin{tabular}[c]{@{}l@{}}1. Patient survey\\ 2. Structured EHR data\end{tabular} & United States & {\ul https://www.researchallofus.org/} \\ \midrule
The 4CE Consortium & i2b2 tranSMART Foundation & \begin{tabular}[c]{@{}l@{}}1. Daily cases \\ 2. Structured EHR data\end{tabular} & Global & {\ul https://covidclinical.net/data/index.html} \\ \midrule
2020 HRS COVID-19 Project &  University of Michigan & \begin{tabular}[c]{@{}l@{}}1. Patient Survey\\ 2. Structured EHR data\end{tabular} & United States & \multicolumn{1}{l}{{\ul https://hrsdata.isr.umich.edu/data-products/2020-hrs-covid-19-project}} \\ \midrule
European CDC COVID-19 Data Tracker & European CDC & Public health data & European & {\ul https://www.ecdc.europa.eu/en/covid-19/data} \\ \midrule
US CDC COVID-19 Data Tracker & United States CDC & Public health data & United States & {\ul https://data.cdc.gov/browse?tags=covid-19} \\ \bottomrule
\end{tabular}%
}
\end{table*}

% \subsection{Data Modalities}
% Imaging data in the context of COVID-19 often includes radiographic images such as those generated by computed tomography (CT) and X-Ray. This is due to the high clinical value of lung imaging to both support the evidence of COVID-19 lung infection and its severity. Fortunately, the importance of XAI in the context of imaging analysis has been heavily explored in the past decade resulting in a wide range of techniques applicable to the most common tasks including image classification, segmentation, clustering, and synthetic image generation. We summarizes publicly available COVID-19 radiology medical imaging datasets could be utilized in implementing XAI (see Table \ref{tab:imaging}). In this section, we focus on these tasks and the approaches used to elucidate model predictions in the context of AI solutions and COVID-19.

% Non-imaging data modalities include the electronic health records (EHRs) and time-series signals. In this section, we will summarize and discuss the interpretation approaches in EHRs before and during the COVID-19 pandemic (as summarized in Fig. \ref{fig:interpretation_EHR_diagram}). Clinicians and researchers use EHRs to predict adverse clinical events, such as mortality or ICU readmission, and to identify top-ranking clinical features to mitigate negative consequences. Despite limitations and challenges, XAI methods have made great impact on non-imaging data modalities.

\subsection{Data Augmentation}
\begin{figure*}[]
    \centering
    \includegraphics[width=\linewidth]{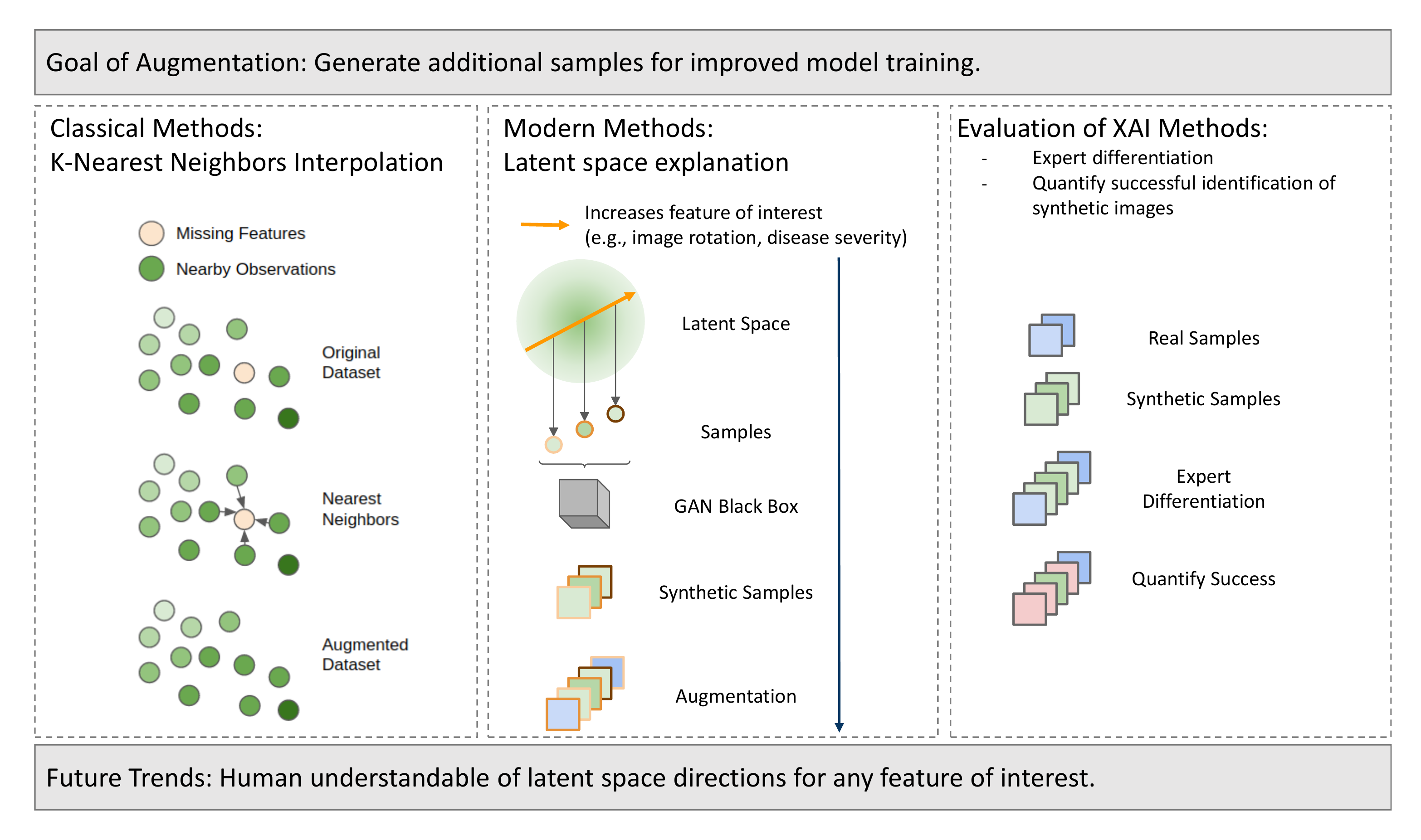}
    \vspace{-8mm}
    \caption{Examples of classical and modern XAI approaches in clinical data augmentation task. It is common to see K-Nearest Neighbors (KNN) interpolation and other classic approaches used in place of more complex modern solutions when performing data augmentation. Due to a dearth of real-world examples of their successes and failures, the modern approach may result in data bias that is difficult to comprehend. The trend in data augmentation has been to increase the number of factors considered and the complexity of data transformations in order to more accurately model the underlying distribution of existing samples.}
    \label{fig:augmentation}
\end{figure*}

The need for labeled data for model training was highlighted in the early stages of the COVID-19 pandemic. This was also a point in time where AI-based solutions could have made the most impact by supplementing scarce public datasets. Future pandemics will likely result in the same urgency for labeled data, and AI-solutions would greatly benefit from synthetic data augmentation. Generative Adversarial Networks (GAN) are used to supplement available labeled COVID radiology data with synthetic images and labels. This allows for improved model training with limited labeled datasets by increasing the number of labeled images available for training. Example of classical and modern data augmentation approaches with model interpretation is shown in Fig. \ref{fig:augmentation}.

Singh et al. tested a wide variety of GAN models to generate synthetic X-ray images while training a COVID-19 detection deep learning model named COVIDscreen \cite{Singh2021-qg}. They compared the quality of four different GAN-based X-ray image generators including Wasserstein GAN (WGAN), least squares GAN (LSGAN), auxiliary classifier GAN (ACGAN), and deep convolutional GAN. They visualized the resulting synthetic X-ray images and showed that WGAN produces visibly higher quality images than the tested alternatives. To the best of our knowledge, this was the first publication to show successful X-ray image generation for COVID-19 data augmentation. A significant limitation of this study was that, although they generated realistic X-ray images using WGAN, they did not leverage this additional data to improve their classifier performance. This is likely due to the lack of label generation during image synthesis which prevents the use of their synthetic images for supervised learning approaches. Despite this limitation, their success in generating synthetic clinical images from a limited COVID-19 dataset illustrated the feasibility of this approach for future work. 

Waheed et al. train an Auxiliary Classifier Generative Adversarial Network (ACGAN) to generate synthetic X-ray images \cite{Waheed2020-fo}. ACGANS take both a label and noise as input to generate new images with known labels. Using COVID-19 status as the label, the proposed model CovidGAN is able to generate normal and COVID-19 images. They train a convolutional neural network (CNN) COVID-19 classifier and compare its performance when trained on a real labeled dataset and a dataset augmented with synthetic images from CovidGAN. They demonstrate that augmentation of their labeled dataset with synthetic images improves classifier performance from 85\% to 95\% classification accuracy.

Loey et al. \cite{Loey2020-jb} trained four CNN classifiers to detect COVID-19 within chest CT images. Synthetic CT images were generated with a conditional GAN (CGAN). They compared the performance of each classifier when trained with four different datasets. Training datasets include: the original dataset alone, the original with morphological augmentation, the original with synthetic images, and the original with morphological augmented combined with synthetic images. They showed that the best classifier (ResNet50) was trained with the original dataset with morphological augmentation (82.64\% balanced accuracy).

Although GANs are widely used for clinical image generation, XAI techniques are not commonly used to understand how they generate the final images from the latent space. Without XAI, it is difficult to detect potential biases in generated images. This is especially important when models are trained on small clinical datasets and subject to a wide range of confounding variables (e.g. hospital-specific signal properties associated with COVID-19 diagnosis). The following novel XAI techniques allow for the interpretation of the GAN latent space in order to understand how sampling of the latent space affects the final image. 

Voynov and Babenko \cite{Voynov2020-kz} created a GAN learning scheme to maximize the interpretability of the GAN latent space. This approach allows the latent space to describe a set of independent image transformations. They showed that this latent space can be visually interpreted and manipulated to generate synthetic images with specific properties (e.g. object rotation, background blur, zoom, etc.). Their method produced synthetic images with interpretable latent space sampling effects across a wide range of datasets including MNIST \cite{noauthor_undated-zz}, AnimeFaces \cite{Jin_undated-yd}, CelebA-HQ \cite{Liu_undated-yv}, and BigGAN \cite{Brock2019-nt} generated images. They show that their interpretation of the latent space can be used to create images with specific properties including zoom, background blur, hair type, skin type, glasses, and many others. These properties were specific to the dataset the GANs were trained on. 

Härkönen et al. \cite{Harkonen2020-ch} also sought to utilize the GAN latent space for image synthesis with specific properties. Instead of re-training models to isolate latent space axis of greatest interpretation, they take existing GANs and identify explainable latent space axes. This allowed them to take an image and change its properties (e.g. convert concrete to grass and object color). The interpretable latent space axes were extracted using principal components analysis (PCA), which requires no additional model training. This technique could also be used to alter image properties (e.g. add wrinkles and white hair to a person) while inheriting the label of the original image. This approach allows synthesis of additional labeled images with known object properties.  

GANs for generating additional radiology images can be interpreted to identify directions of greatest interpretability. This would allow users to understand which image properties the generator is trained to reproduce. There is also the potential to identify latent space directions which significantly correlates with COVID-19 infection presence. Examining these vectors could allow for a better understanding of COVID-19 disease pathology. Non-COVID-19 directions can be used to alter labeled images without affecting class labels which would allow for dataset augmentation with interpretable noise. This image augmentation could improve classifier performance by training it on a wider range of images, reducing the potential for overfitting.

The latent space of COVID-19 GANs are not being examined enough for interpretable features. This is a missed opportunity to identify novel COVID-specific image properties. Using XAI to understand latent space effects on image generation would also allow generation of images with desired properties. XAI also allows examination of image transformation “directions” (e.g. object rotation, zoom) to ensure that they are independent, and not correlated with potential sources of confounding (e.g. scanner model, hospital source, and technician bias). In future pandemics, reliable and explainable synthetic data augmentation approaches may facilitate the training of high-performing AI models to help in the clinical arena.

In addition to data augmentation, synthetic examples can be used to improve model robustness to outliers. Rahman et al. showed that many COVID-19 diagnostic models are vulnerable to attacks by adversarial examples \cite{rahmanAdversarial2021}. Palatnik de Sousa et al. \cite{palatnik_de_sousa_explainable_2021} also demonstrated the utility of adding random ``color artifac'' artifacts to CT images to identify model architecture which are most robust to such perturbation. This illustrates the importance of robust validation of models prior to their integration within clinical settings. XAI may also be used to verify the validity of models' approach to guard against such unexpected, and potentially harmful, results.

\subsection{Outcome Prediction}
\begin{figure*}[]
    \centering
    \includegraphics[width=\linewidth]{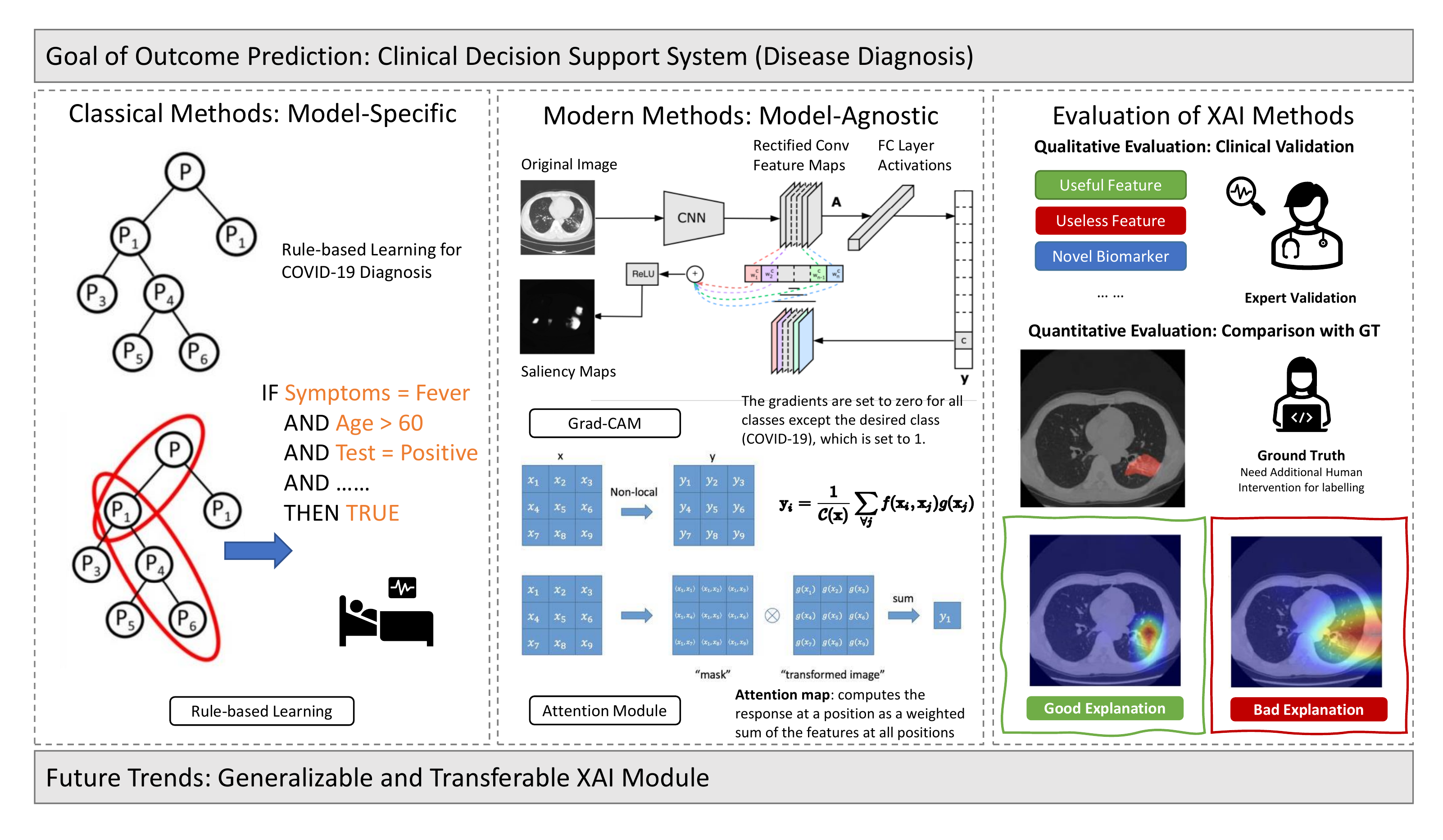}
    \vspace{-5mm}
    \caption{Examples of classical and modern XAI approaches in clinical decision support system. For the task of disease diagnosis, the trend has been to create visualizations of input importance that can be used with a wide variety of popular deep learning models (model-agnostic). This is in contrast to early XAI approaches, which emphasized model-specific solutions in order to improve interpretability.}
    \label{fig:diagnosis}
\end{figure*}

Due to their rapid acquisition times and accessibility, imaging modalities such as X-rays and CT scans have aided clinicians tremendously in diagnosing COVID-19. Radiographic signs, such as airspace opacity, ground-glass opacity, and subsequent consolidation, aid in the diagnosis of COVID-19. However, medical images contain hundreds of slices making diagnosis difficult for clinicians. COVID-19 also exhibits similarities to a variety of other types of pneumonia, posing an additional challenge for clinicians. Although AI-based clinical decision support systems outperform conventional shallow models that have been adapted for clinical use, clinicians frequently lack trust in or understanding of them due to unknown risks, posing a significant barrier to widespread adoption. Thus, XAI-assisted diagnosis via radiological imaging is highly desirable, as it can be viewed as an explainable image classification task for distinguishing COVID-19 from other pneumonia and healthy subjects, as shown in Fig. \ref{fig:diagnosis}.

\begin{figure*}[]
    \centering
    \includegraphics[width=\linewidth]{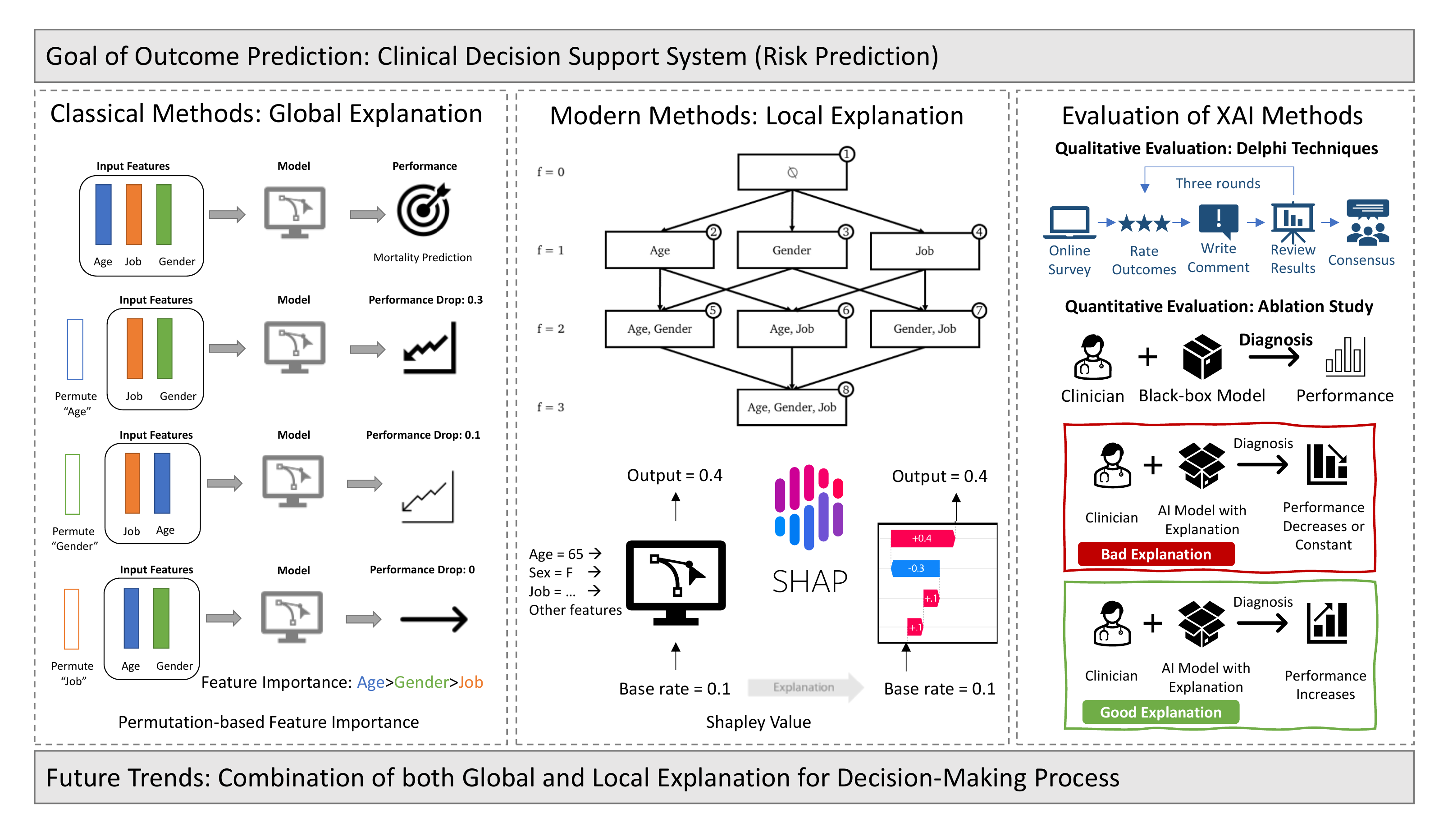}
    \vspace{-5mm}
    \caption{Examples of classical and modern XAI approaches in clinical decision support system. In risk prediction, XAI has increasingly been used to generate local explanations or combination of local and global explanations. This is especially beneficial in the clinical setting, where precision medicine is becoming the standard and patient-specific explanations for risk scores are critical.}
    \label{fig:risk}
\end{figure*}

Another important clinical application in outcome predication task is risk prediction. Clinicians and researchers use Electronic Health Records (EHRs) to predict risk of adverse clinical events, such as mortality or ICU readmission, and to identify top-ranking clinical features to mitigate negative consequences (see Fig. \ref{fig:risk}).

Interpretation by feature scoring, also known as saliency, relevance, or feature attribution, is the most common XAI strategy in outcome prediction.
Interpretation by feature scoring finds evidence supporting individual predictions by calculating importance scores associated with each feature of the input. 
Specifically, given an input, we need to find a vector of importance scores that is the same size as the input. 
In general, feature scoring can be grouped into five categories: perturbation-based, activation-based, gradient-based, mixed-based (combination of activation-based and gradient-based), and attention-based approaches.

\subsubsection{Perturbation-based Approach}
Perturbation is the simplest way to analyze the effect of changing the input features on the output of an AI model. 
This can be implemented by removing, masking, or modifying certain input features, running the forward pass, and then measuring the difference from the original output. 
The input features affecting the output the most are ranked as the most important features. 

Permutation- or occlusion-based methods measure the importance of a feature by calculating the increase in the model’s prediction error after permuting the feature. A feature occlusion study \cite{zeiler2014visualizing} was performed to show the influence of occluding regions of the input image to the confidence score predicted by the CNN model. The occlusion map was computed by replacing a small area of the image with a pure white patch and generating a prediction on the occluded image. While systematically sliding the white patch across the whole image, the prediction score on the occluded image was recorded as an individual pixel of the corresponding occlusion map. In biomedical application, Tang et al. \cite{tang2019interpretable} utilized occlusion mapping to demonstrate that networks learn patterns agreeing with accepted pathological features in Alzheimer’s disease. 

In the COVID-19 imaging applications, Gomes et al. \cite{gomes_features_2021} presented an interpretable method for extracting semantic features from X-rays images that correlate to severity from a data set with patient ICU admission labels. The interpretable results pointed out that a few features explain most of the variance between patients admitted in ICU. To properly handle the limited data sets, a state-of-the-art lung segmentation network was also trained and presented, together with the use of low-complexity and interpretable models to avoid overfitting. Casiraghi et al. \cite{casiraghiExplainable2020} calculated COVID-19 patient risk for significant complications from radiographic features extracted using deep learning and non-imaging features. Random Forest (RF) and Boruta feature selection were used for feature selection. The most important features were then used to train a final RF model to predict risk. In order to maximize final model interpretability, they generated a sequence of steps to generate an association decision tree from the final RF model. The final association tree is easily interpretable by experts. 

Another perturbation-based approach is Shapley value sampling \cite{vstrumbelj2014explaining}, which estimates input feature importance via sampling and re-running the model. Calculating these Shapley feature importance values is computationally expensive as the network has to be run for each sample and feature (sample $\times$ number of features) times. Lunberg et al. \cite{lundberg2017unified} proposed a fast implementation for tree-based models named SHapley Additive exPlanation (SHAP) to boost the calculation process. Shapley values can be calculated to discover how to divide the payoff equitably by treating the input features as participants in a coalition game. SHAP has been shown to be helpful in explaining clinical decision-making in the medical field both from image \cite{singh2020interpretation} and non-image \cite{lundberg2018explainable} inputs and has also been well explored under COVID-19 cases \cite{zoabi2020covid,wu2020ai,ong_comparative_2021}.

Similarly, Local Interpretable Model-Agnostic Explanations (LIME) \cite{ribeiro2016should} is a procedure that enables an understanding of how the input features of a deep learning model affect its predictions. For instance, LIME determines the set of super-pixels (a patch of pixels) that have the most grounded relationship with a prediction label when used for image classification. LIME performs clarifications by creating a new dataset of random perturbations (each with its own forecast) around the occasion and then fitting a weighted neighborhood proxy model. Typically, this neighborhood model is a simpler one with natural interpretability, such as a linear regression model. LIME generates perturbations by turning on and off a subset of the super-pixels in the image. 
To generate a human-readable representation, LIME attempts to determine the importance of contiguous superpixels in a source image relative to the output class. It has been widely implemented in COVID-19 diagnosis tasks \cite{ahsan2020study, ong_comparative_2021, ahsan_detection_2021, ye_explainable_2021} to further explain the process of feature extraction, which contributes to a better understanding of what features in CT/X-ray images characterize the onset of COVID-19.
Ahsan et al. \cite{ahsan2020study} implemented LIME to interpret top features in COVID-19 X-ray imaging and build trust in an AI framework to distinguish between patients with COVID-19 symptoms with other patients.
Similarly, Ong et al. \cite{ong_comparative_2021} implemented both SHAP and LIME to expound and interpret how Squeezenet performs COVID-19 classification and highlight the area of interest where they can help to increase the transparency and the interpretability of the deep model.

\subsubsection{Activation-based Approach} 
The group of activation-based approaches can identify important regions in a forward pass by obtaining or approximating the activations of intermediate variables in a DL model. 
Because extracted features within deep layers are closer to the classification layer, they capture more class-discriminative information than those in bottom layers. 
Erhan et al. \cite{erhan2010understanding} concentrated on input patterns that maximize the activation of a particular hidden unit, called Activation Maximization, to illustrate the relevance of features in DL models. 
Zhou et al. \cite{zhou2016learning} proposed Class Activation Maps (CAM), which used global average pooling to calculate the spatial average of feature maps in the last convolutional layer of a CNN. Han et al. \cite{han2020accurate} proposed an attention-based deep 3D multiple instance learning (AD3D-MIL) to semantically generate deep 3D instance following the potential infection regions. Additionally, AD3D-MIL used an attention-based pooling to gain insight into each instance's contribution over a broader spectrum, allowing for more in-depth analysis. In comparison to conventional CAM, AD3D-MIL was capable of precisely detecting COVID-19 infection regions via key instances in 3D models. It achieved an accurate and interpretable COVID-19 screening that has the potential to be generalized to large-scale screening in clinical practice. 

\subsubsection{Gradient-based Approach} % ReLu, saliency, LRP

\begin{figure}[]
    \centering
    \includegraphics[width=\linewidth]{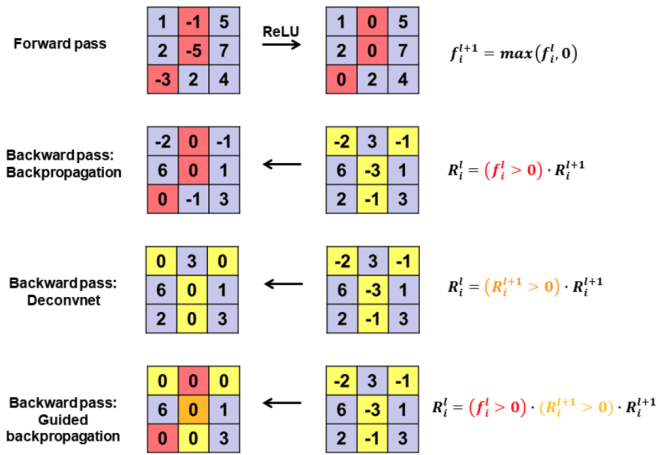}
    \vspace{-5mm}
    \caption{Comparisons of different methods propagating through ReLU: forward pass, back-propagation, deconvnet, and guided back-propagation. Adapted from \cite{springenberg2014striving}.}
    \label{fig:RELU}
\end{figure}

Gradient-based approaches identify important features by evaluating gradients of an input through back-propagation. The intuition behind this idea is that input features with large gradients have the largest effects on predictions. Simonyan et al. \cite{simonyan2013deep} constructed the importance map of input features by calculating the absolute value of partial derivatives of class score with respect to the input through back-propagation. However, feature importance calculated above could be noisy because of the saturation problems caused by the existence of non-linear operations such as rectified linear units (ReLU). That is, changes in gradients could be removed in a backward pass if the input to ReLU are negative. To address this issue, several modifications to the way ReLU is handled have been proposed, as illustrated in Fig. \ref{fig:RELU}. Springenberg et al. \cite{springenberg2014striving} proposed guided back-propagation by combining standard back-propagation and the “deconvnet” approach: keep gradients only when both bottom input and top gradients are positive. Thus, guided back-propagation can sharpen the feature importance scores compared to vanilla gradients by back-propagation. 

Layer-wise Relevance Propagation (LRP) proposed by Bach et al. \cite{bach2015pixel} is also used to find relevance scores for individual features in the input data by decomposing the output predictions of the DL models. 
The relevance score for each input feature is derived by back-propagating the output class node's class scores towards the input layer. The propagation is governed by a stringent conservation property, which requires an equal redistribution of the relevance received by a neuron. LRP was used in COVID-19 X-ray imaging to offer reasons for diagnostic predictions and to pinpoint crucial spots on the patients' chests. \cite{karim2020deepcovidexplainer, bassi2021deep}. 

Saliency map generation and analysis was first introduced by Simonyan et al. \cite{simonyan2013deep} to calculate the pixel importance. The importance score of each pixel is generated using the gradient of the output class category relative to an input picture, and a meaningful summary of pixel importance can be obtained by examining which positive gradients had the most effect on the output. Shamout et al. \cite{shamout2021artificial} proposed a a data-driven approach for automatic prediction of deterioration risk using a deep neural network that learns from chest X-ray images and a gradient boosting model that learns from routine clinical variables. To illustrate the interpretability of proposed model, they performed the saliency maps for all time windows (24, 48, 72, and 96 h) to highlight regions that contain visual patterns such as airspace opacities and consolidation, which are correlated with clinical deterioration. These saliency maps could be used to guide the extraction of six regions of interest patches from the entire image, each of which is then assigned a score indicating its relevance to the prediction task. Similarly, \cite{wu2021jcs, barbosa2021machine, qian2020m, singh2021interpretable, singh_screening_2021} also include saliency maps as an explainable deliverable to interpret deep models and find potential infection regions in COVID-19 diagnosis and detection.

\subsubsection{Mixed-based Approach} % gradcam, gradcam++
Both activation-based and gradient-based methods have their own set of benefits and drawbacks. Specifically, activation-based methods generate feature scores that are more class discriminative, but they suffer from the coarse resolution of importance scores. On the other hand, although gradient-based methods produce fine resolution of feature scores, they tend not to show ability to differentiate between classes. Gradient-based and activation-based approaches could be combined to produce both fine and discriminative features importance scores.

\begin{figure}[]
    \centering
    \includegraphics[width=\linewidth]{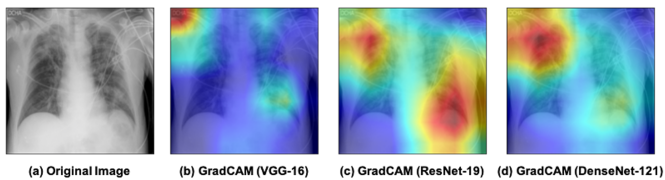}
    \vspace{-5mm}
    \caption{Comparison of different network visualization with GradCAM. The heatmaps are overlapped on the original image, where the red color highlights the activation region associated with the predicted class (COVID-19). (a) original image (b) GradCAM with VGG-16 (c) GradCAM with ResNet-18 (d) GradCAM with DenseNet 121.}
    \label{fig:casestudy}
\end{figure}

Gradient-weighted Class Activation Mapping (Grad-CAM) \cite{selvaraju2017grad} proposed by Selvaraju et al. uses the gradients flowing down to the last convolutional layer to multiply CAM from a forward pass. The resolution is enhanced by multiplying Grad-CAM with guided-backpropagated gradients. Class-specific queries and counterfactual explanations supported by Grad-CAM enable the visualization of portions of a picture that have a detrimental impact on model output, as shown in Fig. \ref{fig:casestudy}. Grad-CAM++ \cite{chattopadhay2018grad} replaces the globally averaged gradients in Grad-CAM with a weighted average of the pixel-wise gradients since the weights of pixels contribute to the final prediction, which leads to better visual explanations of CNN model predictions. It addresses the shortcomings of Grad-CAM, especially multiple occurrences of a class in an image and poor object localization. 

Due to the vanishing non-linearity of classifiers, CAM is often unsuitable for interpreting deep learning models in COVID-19 image classification tasks. Grad-CAM and Grad-CAM++ both enhanced the CAM procedure to enable better visualizations of deeper CNN models and are usually considered the most popular interpretation strategy in COVID-19 automatic diagnosis on radiographic imaging \cite{li2020using, wang2021psspnn, song2021deep, arias2020artificial, brunese2020explainable, panwar2020deep, alshazly2021explainable}. Additionally, Oh et al. \cite{oh2020deep} proposed patch-wise deep learning architecture to investigate potential biomarkers in X-ray images and find the globally distributed localized intensity variation, which can be a discriminatory feature for COVID-19. They extended the idea of Grad-CAM to a novel probabilistic Grad-CAM that took patch-wise disease probability into account, resulting in more precise interpretable saliency maps that are strongly correlated with radiological findings. 

\subsubsection{Attention-based Approach}
%To help the model make more reasonable classifications and generate interpretable results, researchers borrowed an idea from the attention mechanism, which is widely used in artificial intelligence applications. 
Attention mechanism is a critical component of human perception, as it enables humans to selectively focus on critical portions of an image rather than processing the entire scene. Simulating the human visual system's selective attention mechanism is also critical for comprehending the mechanisms underlying black-box neural networks. Attention mechanism has been widely applied to computer vision applications \cite{wang2018non}, endowing the model with several new characteristics: 1) determine which portion of the inputs to focus on; 2) allocate limited computing resources to more critical components. 

The effectiveness of attention mechanism has been demonstrated in a variety of medical image analysis tasks. Specifically, several state-of-the-art methods have been proposed to leverage attention mechanisms in order to improve the discriminative capability of classification models for both X-ray \cite{chen2019lesion} and CT image analysis tasks \cite{xu2019pulmonary,chikontwe_dual_2021}. In the application of COVID-19 diagnosis, Shi et al. \cite{shi_covid-19_2021} propose an explainable attention-transfer classification model based on a knowledge distillation network structure to address the difficulties associated with automatically differentiating COVID-19 and community-acquired pneumonia from healthy lungs in radiographic imaging. Extensive experiments on public radiographic datasets demonstrated the explainability of the proposed attention module in diagnosing COVID-19.

In addition to medical imaging, the attention mechanism is also useful in other feature interpretation setting, such as unstructured clinical notes with natural language processing (NLP). Diagnostic coding of clinical notes is a task that aims to provide patients with a coded summary of their disease-related information. Recently, Dong et al. \cite{dong2021explainable} proposed a novel Hierarchical Label-wise Attention Network (HLAN) to automate such a medical coding process and to interpret model prediction results by evaluating the attention weights at word and sentence level. The label-wise attention scores in the proposed HLAN model provide comprehensive and robust explanation to support the prediction. Zhang et al. \cite{zhang2018patient2vec} proposed Patient2Vec to learn interpretable deep representations and predict risk of hospitalization on EHR data. The backbones of the model are gated recurrent units (GRU) and a hierarchical attention mechanism that learn and interpret the importance of clinical events on individual patients. 

Recurrent neural network (RNN)-based variants with attention modules have been widely explored for severity assessment in EHRs \cite{choi2016retain, kwon2018retainvis, kaji2019attention, shickel2019deepsofa, yin2019domain, li2020marrying, chen2020interpretable}. Choi et al \cite{choi2016retain} proposed a reversed time attention model (RETAIN) to learn interpretable representation on EHRs. RETAIN is based on a two-level neural attention model that detects significant clinical variables in influential hospital visits in reverse time order. % Different from standard attention mechanisms in NLP, RETAIN recovers the sequential information in EHR data using RNNs and preserves interpretability using a multi-layer perceptron (MLP) on embedded clinical visit information. 
 Kaji et al. \cite{kaji2019attention} combined RNNs with a variable-level attention mechanism to interpret the model and results at the level of input variables.
 Shickel et al. \cite{shickel2019deepsofa} proposed an interpretable illness severity scoring framework named DeepSOFA that can predict the risk of in-hospital mortality at any time point during the ICU stay. By incorporating the self-attention mechanism, the RNN variant with gated recurrent unit (GRU) analyzes hourly-based time-series clinical data in the ICU. The self-attention module in the recurrent neural network is designed to highlight certain time steps of the input data which the model perceives to be most important in mortality prediction. 
 Yin et al. \cite{yin2019domain} developed a domain-knowledge–guided recurrent neural network, an interpretable RNN model with a graph-based attention mechanism that incorporates clinical domain knowledge learnt from a public clinical knowledge graph. Li et al. \cite{li2020marrying} extended the proposed method to enable the exploration and interpretation on clinical risk prediction tasks through visualization and interaction of deep learning models.

In addition to RNN variants, Lauritsen et al. \cite{lauritsen2020explainable} proposed an explainable AI Early Warning Score (xAI-EWS) to predict acute critical illness on EHRs. The proposed model consists of a Temporal Convolution Network (TCN) prediction module and a Deep Taylor Decomposition (DTD) explanation module. Through computing back-propagated relevance scores, the DTD module identifies relevant clinical parameters at a given time for a prediction generated by the TCN. The xAI-EWS enables explanations in real time and allows the physicians to understand which clinical variables or parameters cause high EWS scores or changes in the EWS score.

However, the attention mechanism continues to struggle when confronted with missing coding, rare labels, or clinical notes containing subtle errors. Additionally, clinical notes in real-world clinical practice frequently contain multiple sentences, and it is unknown how well the attention mechanism would function when interpreting multiple sentences. Additionally, external domain knowledge in the medical field is required to verify interpretation results. In general, the attention mechanism has enormous potential for emphasizing critical features and fostering trust in clinical practice.

\begin{figure*}[]
    \centering
    \includegraphics[width=\linewidth]{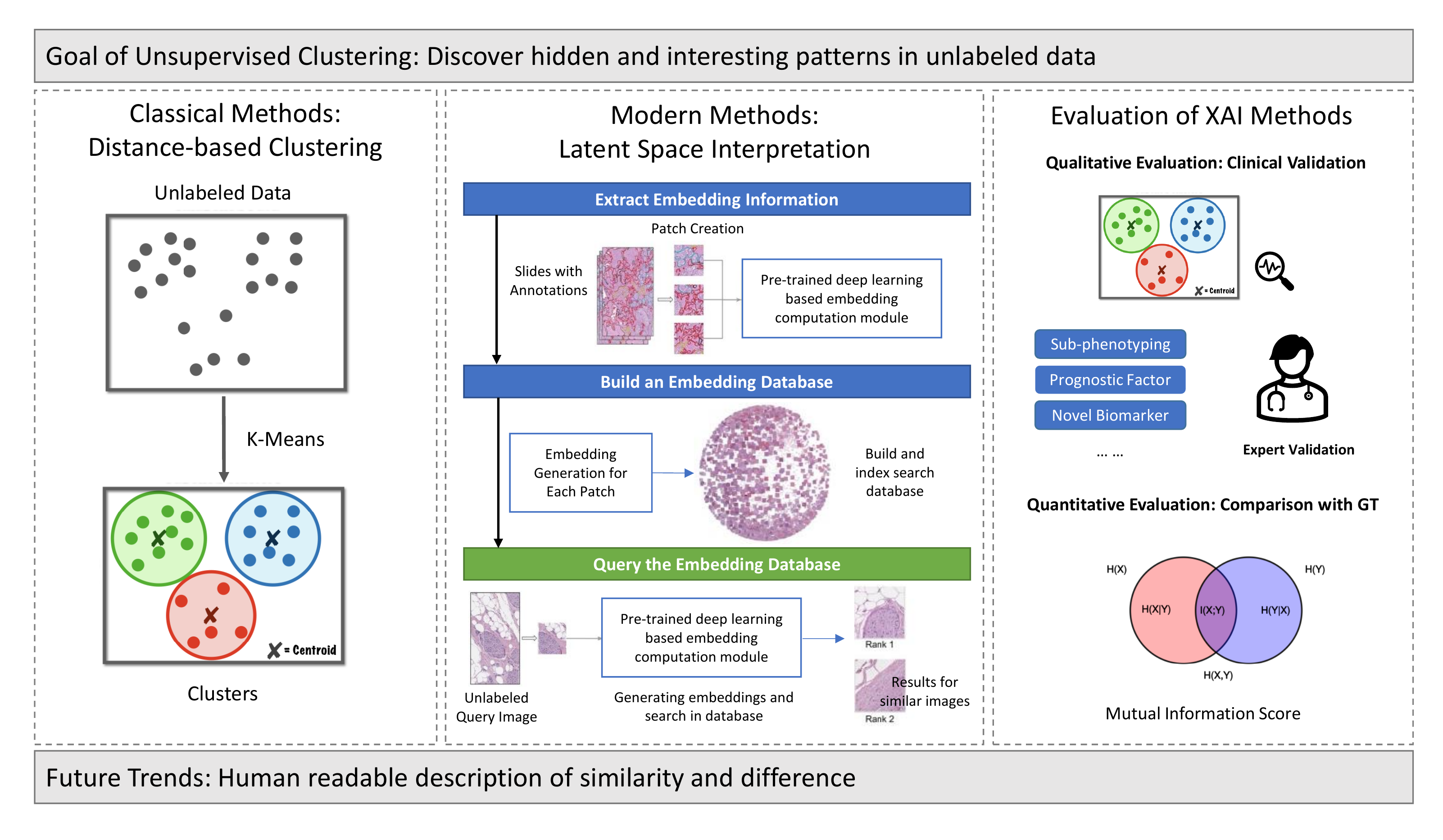}
    \vspace{-8mm}
    \caption{Examples of classical and modern XAI approaches in unsupervised clustering task. Unsupervised clustering has benefited from the use of latent spaces generated by deep learning models to generate sample similarities. This shift from the conventional approach of calculating input feature distances enables the use of custom transformations to optimize the space in which similarity is measured. This can result in improved sample disentanglement. The histopathology images are adapted from \cite{hegde2019similar}.}
    \label{fig:clustering}
\end{figure*}

\subsection{Unsupervised Clustering}
Development of an AI-based diagnosis system for COVID-19 was different from traditional epidemiological challenges: in the early stage of a new disease there is limited amounts of available data, especially diagnostic information \cite{shi2020review}. The major downside of traditional deep learning methods is that they largely rely on the availability of labeled data, while COVID-19 datasets often contain incomplete or inaccurate labels. In biomedical applications, unsupervised learning has the benefit of not needing labeled data to still train, extract features, and cluster data, which makes it a great candidate for COVID-19 diagnosis (see Fig. \ref{fig:clustering}). 

The application of unsupervised learning approaches, especially clustering techniques, represents a powerful means of data exploration. Discovering underlying data characteristics, grouping similar measurements together, and identifying patterns of interest are some of the applications which can be tackled through clustering. Being unsupervised, clustering does not always provide clear and precise insight into the produced output, especially when the input data structure and distribution are complex and unlabeled. Applying XAI can allow researchers to understand the reasons leading to a particular decision under clinical scenarios and suggest an explanation to the clustering results for the end-users.

Recent advances in Auto-Encoders (AEs) have shown their ability to learn strong feature representations for image clustering \cite{song2013auto,lim2020deep,prasad2020variational}. 
By designing the constraint of the distance between data and cluster centers well, Song et al. \cite{song2013auto} artificially re-aligned each point in the latent space of an AE to its nearest class neighbors during training to obtain a stable and compact representation suitable for clustering. 
Lim et al. \cite{lim2020deep} generalize Song's approach by introducing a Bayesian Gaussian mixture model for clustering in the latent space and replacing the input points with probability distributions which can better capture more hidden variables and hyperparameters. 
Prasad et al. \cite{prasad2020variational} introduced a Gaussian Mixture prior to help clustering based on Variational Auto-Encoders to efficiently learn data distribution and discriminate between different clusters in a latent space.

In addition to guided feature representation achieved by AEs, King et al. \cite{king2020unsupervised} applied chest X-ray images of COVID-19 patients to a Self-Organizing Feature Map (SOFM) and found a distinct classification between COVID-19 and healthy patients. SOFM was first proposed to provide data visualization to cluster unlabeled X-ray images as well as reducing the dimensions of data to a map to understand high dimensional data. SOFM applied competitive learning to selectively tune the output neurons to the classes of the input patterns and then cluster their weights in locations respective to each other based off the feature similarities. They demonstrate that image clustering methods, specifically with SOFM networks, can cluster COVID-19 chest X-ray images and extract their features successfully to generate explainable results. 

Yadav \cite{yadav_lung-gans_2021} proposed a deep unsupervised framework called Lung-GANs to learn interpretable representations of lung disease images using only unlabeled data and classify COVID-19 from chest CT and X-ray images. They extracted the lung features learned by the model to train a support vector machine and a stacking classifier and demonstrated the performance of proposed unsupervised models in lung disease classification. They visualized the features learnt by Lung-GANs to interpret deep models and empirically evaluate its effectiveness in classifying lung diseases.

Singh et al. \cite{singh_these_2021} used image embedding generated from a prototypical part network (ProtoPNet) inspired network to calculate similarities and differences of X-ray image patches to known examples of pathology and healthy patches. This metric was then used to classify subjects into COVID-19 positive, pneumonia, or healthy classes.

The task of image clustering in COVID-19 and other clinical scenarios naturally requires good feature representation to capture the distribution of the data and subsequently differentiate one category from one another.
In general, unsupervised clustering is an XAI technique which can be implemented to validate that images cluster in meaningful groups and facilitate expert annotation by extrapolating labels within samples belonging to the same cluster, when labels need be estimated.

\subsection{Image Segmentation}
\begin{figure*}[]
    \centering
    \includegraphics[width=\linewidth]{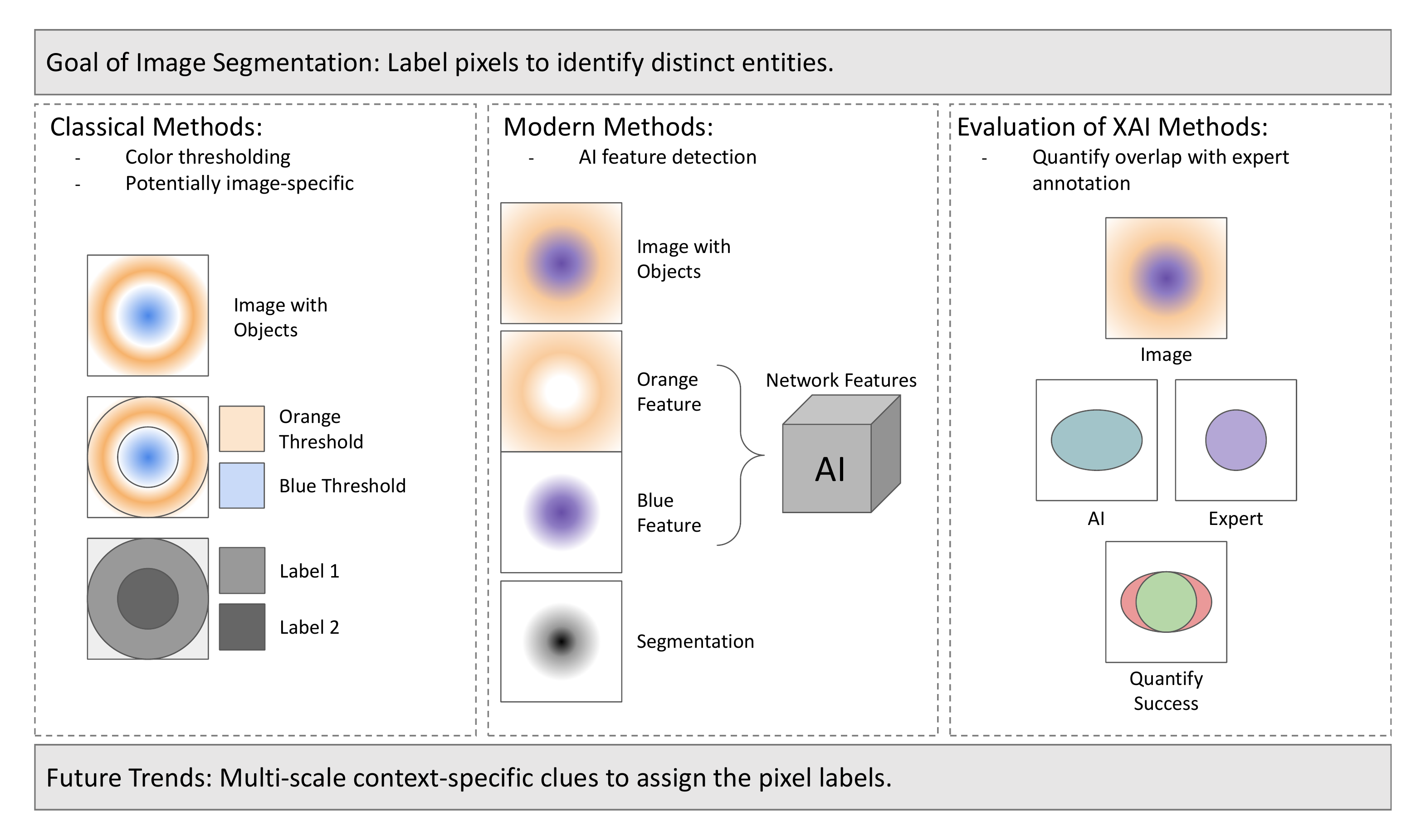}
    \vspace{-8mm}
    \caption{Examples of classical and modern XAI approaches in image segmentation task. Segmentation models have progressed from being highly interpretable (when simple color thresholds are used) to requiring numerous nonlinear transformations to generate the final segmentation. Although XAI approaches to image segmentation are not widely used, recent techniques have used the model activation maps generated by deep layers to identify significant associations with the final segmentation.}
    \label{fig:segmentation}
\end{figure*}

Segmentation algorithms make pixel-level classifications of images and the overall segmentation produced provide insight into the decisions of the model. In the realm of XAI, image segmentation itself can be considered highly interpretable. Therefore, explanations of the segmentation process are currently not widely explored for medical image analysis. In the current climate, segmentation algorithms function as useful tools for isolating regions significant to COVID-19 diagnosis or for determining infection severity. Application of explainable AI techniques to segmentation techniques could provide valuable information to improve COVID-19 segmentation approaches, as shown in Fig. \ref{fig:segmentation}. 

Current COVID-19 segmentation approaches often use convolutional neural networks to delineate the regions of interest. One example of this model was developed by Saeedizadeh et al. for segmenting CT images of COVID-19 patients based on U-Net, which they call TV-UNet \cite{saeedizadeh2020covid}. The framework was trained to detect ground glass regions on the pixel level, which are indicative of infected regions, and to segment them from normal tissue. TV-UNet differs from regular U-Net by the addition of an explicit regularization term in the training loss function which the authors report improves connectivity for predicted segmentations. Their model was trained on a COVID CT segmentation dataset with three different types of ground truth masks and reported an average DICE coefficient score of 0.864 and an average precision of 0.94. However, the results of the segmentation algorithm do not provide any intuition on why the model made the decisions it did. Part of this is due to the black box nature of U-Net. The residual connections between layers are inherently obscure to human intuition which makes it difficult to understand how U-Net decided to apply the labels. Application of a technique that explains the model’s decision-making process could provide information on possible biases in the model and ways to improve it. Pennisi et al. \cite{pennisiAn2021} achieved sensitivity and specificity of COVID-19 lesion categorization of over 90\% using a combination of lung lobe segmentation followed by lesion classification. In addition, they also created a clinician-facing user interface to visualize model prediction. This expert oversight was leveraged to improve future prediction by integrating clinician feedback through the same user interface (expert in the loop). Wang et al.\cite{wang_joint_2021} proposed an interpretable DeepSC-COVID designed with 3 subnets: a cross-task feature subnet for feature extraction, a 3D lesion subnet for lesion segmentation, and a classification subnet for disease diagnosis. Different from the single-scale self-attention constrained mechanism \cite{wang2018non}, they implemented multi-scale attention constraint to generate more fine-grained visualization maps for potential infections. 

Image morphology-based segmentation approaches are not as common within the context of COVID-19 image segmentation, but they do exist. An example from \cite{yao_csgbbnet_2021} demonstrates the successful use of an maximum entropy threshold segmentation-based method along with fundamental image processing techniques, such as erosion and dilation, to isolate a final lung-only binary mask. These lung masks can also be used to generate bounding boxes to limit classification to regions surrounding, and including, lung tissue \cite{nedumkunnel_explainable_2021}.
In addition to lesion segmentation, some approaches first segment lung tissue prior to classification or further segmentation of clinically-relevant lesions. Jadhav et al. utilized this approach to allow radiologists to use a user-interface to view the two and three-dimensional CT regions used for the classification task with a saliency map overlay \cite{jadhav_covid-view_2021}. This combination of XAI approaches sought to increase radiologist trust of classification predictions by gaining multiple visual insights of the automated workflow.

Natekar et al. described one such method of explaining segmentation algorithms known as network dissection \cite{natekar2020demystifiying}. Their focus was on explaining segmentations done on MR images of brain tumors with U-Net, but the techniques could be applicable to COVID-19 segmentation. They explain network dissection as follows: for a single filter in a single layer, collect the activation maps of all input images and determine the pixel-level distribution over the entire dataset. In CNNs, individual filters can focus on learning specific areas or features in an image, however, it is not clear from the outside which filter does which. Dissecting the network would make the purpose of each filter clearer and allow for better understanding of the decisions made by the model. Application to COVID-19 algorithms such as TV-UNet could allow for visualization of specific features that the model looks for to make a segmentation decision, thereby increasing user confidence in the model.

Another COVID-19 segmentation approach is the joint classification and segmentation diagnosis system developed by Wu et al. \cite{wu2021jcs}. In their framework, they include an explainable classification model and segmentation model that work together to provide diagnosis prediction for COVID-19. Their segmentation is done via an encoder-decoder architecture based on VGG-16, plus the addition of an Enhanced Feature Module to the encoder which the authors proposed to improve the extracted feature maps. They trained and tested their model on a private COVID dataset and reported a DICE coefficient score of 0.783. Typically, image segmentation tasks are used to help explain classification decisions but the authors of this paper extend this idea by having the classification also help explain the segmentation. The segmentation algorithm references information from the classifier by merging their feature maps together to improve its decisions but this also helps indicate the reasoning behind the decisions made when producing segmentation. Utilizing classification information to help train and explain segmentation is an avenue which merits further exploration.
\section{Explanation Representation}
\label{sec:representation}
\subsection{Heat-map Visualization}
Visualizing the network is advantageous for diagnosing model problems, interpreting the models' meaning, or simply guiding deep learning concepts. For instance, we can visualize decision boundaries, model weights, activations, and gradients for a CNN model and extracting feature maps from hidden convolutional layers. Heat-maps or saliency maps visualization is the most frequently used method for interpreting convolutional neural network predictions which plays an important role in the model interpretation and explanation representation.
 
In computer vision, heat-map or saliency map is an image that highlights important regions to focus user attention. The purpose of a saliency map is to show how important a pixel is to the human visual system. Visualization provides an interpretable technique to investigate hidden layers in deep models. For example, in a COVID-19 imaging classification task, saliency maps generated by gradient-based XAI methods are widely used to measure the spatial support of a particular class in each image \cite{shamout2021artificial,wu2021jcs, barbosa2021machine, qian2020m,singh2021interpretable,singh_screening_2021,li2020using, wang2021psspnn, song2021deep, arias2020artificial, brunese2020explainable, panwar2020deep, alshazly2021explainable,oh2020deep,chikontwe_dual_2021,shi_covid-19_2021}. 

% Clinical Applications

\subsection{Decision Tree}
Most existing papers on XAI for COVID-19 EHRs rely on decision tree-based approaches for mortality prediction and ICU readmission prediction \cite{yan2020interpretable, vaid2020machine, wollenstein2020physiological, pan2020prognostic, lu2020explainable, sanchez2020machine, brinati2020detection, alves2021explaining, estiri2021individualized, qomariyah2021tree, famiglini2021prediction}. Before COVID-19 pandemic, decision tree-based models \cite{ou2020rupture, liu2018interpretable, song2020cross, yang2016predicting, misra2021early, che2016interpretable} have already been widely used for interpreting clinical features in EHRs under difference clinical settings.

Gradient boosted tree (XGBoost) models can generate feature importance and consequently identify top-ranking features and potential biomarkers. Yet these models in existing COVID-19 papers are often validated on single-center datasets, and their interpretation for clinical practice is limited. Yan et al. \cite{yan2020interpretable} presented an interpretable XGBoost classifier to predict patient mortality and identify critical predictive biomarkers on medical records of 485 infected patients in the city of Wuhan. The machine learning model achieved over 90\% accuracy in predicting patient mortality more than 10 days in advance. In addition, the author modified the XGBoost classifier into an interpretable “single-tree XGBoost” to identify the top three biomarkers: lactic dehydrogenase, lymphocyte and high-sensitivity C-reactive protein (hs-CRP), which were consistent with medical domain knowledge. Although the author reported great performance on mortality prediction and identifying the top three clinical biomarkers, the model was only validated on a small sample size without external validation. In addition, as 88\% of patients survived and 12\% of patients died, it is inappropriate to use accuracy as the only evaluation performance on this imbalanced dataset. Subsequently, this paper has been challenged by a few follow up articles from different perspectives. 

% Giacobbe et al \cite{giacobbe2020clinical} presented a “possible alternative, non-mutually exclusive explanation of the results” regarding the clinical interpretation. The author argued that very high hs-CRP levels may reflect the development of bacterial superinfection, which could be the true reason of increased mortality. Consequently, the interpretation from the original paper may change the clinical perspective and shift “the interest of clinicians from identifying patients at higher risk of death to identifying those at higher risk of superinfections or other complications”. 

Three other papers \cite{dupuis2020limited, quanjel2020replication, barish2020external} challenged the original paper \cite{yan2020interpretable} on its limited performance of mortality prediction and less applicable clinical interpretation using external datasets. Barish et al. \cite{barish2020external} demonstrated the ineffectiveness of the original model as an admission triage tool on the internal dataset. Furthermore, the original decision-tree-based model is not applicable on an external dataset which was collected from 12 acute care hospitals in New York, USA. Therefore, external validation is a critical step in verifying the proposed model before adopting the identified biomarkers in clinical practice.

Similarly, Alves et al. also used a post-hoc method of explaining a random forest (RF) classifier result using an instance based interpretation \cite{alves2021explaining}. This involves minimizing the distance between a decision tree classifier and the RF model for a specific sample. This approach was called decision tree-based explanation. Additional nearby samples are generated for the purposes of optimization by adding noise to the chosen sample features. This approach allows decision trees to be generated for specific patients which help clinicians understand the model decision. This approach is limited to explaining the decision for individual samples, and does not offer a global explanation of model decision making. Shapley Additive Explanations (SHAP) and LIME were also used to generate additional global and local explanations respectively.

\subsection{Feature Importance}
%Visualization of feature importance \cite{oh2020deep,tang2020prediction,pan2020prognostic,cao2020impact,dong2021explainable,cavallaro2021contrasting,abdulaal2020prognostic,chikontwe_dual_2021,ye_explainable_2021,singh_interpretable_2021,nedumkunnel_explainable_2021}
Under COVID-19 pandemic, an alternative explanation representation approach is to calculate the feature interpretation score after predicting mortality and other critical health events \cite{torres2020pandemyc, vaid2020machine, foieni2020derivation, pan2020prognostic, tang2020prediction, zheng2021interpretable, makridis2021designing, haimovich2020development, abdulaal2020prognostic, cavallaro2021contrasting, qomariyah2021tree, patel2021natural, magunia2021machine, singhal2021eards, nguyen2021budget, rodriguez2021development, wu2021interpretable}. Before the pandemic, clinical scoring systems \cite{martini2021drivers, rodriguez2019interpretation, dai2020development, caicedo2019iseeu} had already been widely used for interpreting clinical features in EHRs. In clincial applications, SHAP \cite{lundberg2017unified} and permutation-based \cite{fisher2019all} feature importance are the most popular ways for feature importance calculation. 

SHAP feature importance \cite{lundberg2017unified} is widely used to explain the prediction of an instance by computing the contribution of each feature to the prediction. The SHAP explanation method computes Shapley values based on coalitional game theory. Pan et al. \cite{pan2020prognostic} identified important clinical features using SHAP score and LIME score on EHR data from 123 COVID-19 ICU patients. Afterwards, the authors built a reliable XGBoost classifier for mortality prediction and ranked selected features. The combination of clinical score and decision tree-based algorithm enables a more generalizable method to identify and interpretable clinical features of top importance. 

In addition to prognostic assessment, SHAP feature importance was also adopted to identify both the city-level and the national-level contributing factors in curbing the spread of COVID-19. The findings may help researchers and policymaker to implement effective responses in mitigating the consequences of the pandemic. Cao et al \cite{cao2020impact} applied XGBoost model to predict new COVID-19 cases and growth rate using six categories of variables, including travel-related factors, medical factors, socioeconomic factors, environmental factors and influenza-like illness factors. By calculating SHAP scores, the author quantitatively evaluated these contributing factors to new cases and growth rate. The author indicated that the population movement from Wuhan and non-Wuhan regions to the target city are both significant factors that contribute to new cases. One major concern with SHAP score is how to interpret the negative SHAP values in this task. Instead of indicating negative impact, negative SHAP values might be interpreted as no impact or compromised effect. More careful consideration is definitely needed. 

Apart from SHAP, Foieni et al. \cite{foieni2020derivation} derived a score to identify the risk of in-hospital mortality and clinical evolution based on the linear regression of 8 clinical and laboratory variables on $119$ COVID-19 patients. These 8 clinical and laboratory variables are significantly associated with model prediction outcomes. One limitation of this work is that the score is defined as a linear combination of 8 variables. Consequently, the score cannot be generalized on external dataset with different clinical and laboratory variables. 

Meanwhile, DeepCOVID-Net \cite{ramchandani2020deepcovidnet} was proposed to forecast new COVID-19 cases using county-level features, such as census data, mobility data and past infection data. Feature importance was estimated in two steps: 1) evaluate model accuracy on a small subset of the training data; 2) loop through all features, independently randomize the values of one feature at a time, and re-calculate the model accuracy on the same training set. Higher importance was assigned to features with lower performance drop during the evaluation process. As such, the author identified the top three features: past rise in infected cases, cumulative cases in all counties, and incoming county-wise mobility. Although these three features are reasonable in common sense, interpreting interactions at the individual feature level might still be difficult, as the model accounts for the higher-order interaction between these feature groups. 

\subsection{Interactive User Interface}
% User interface \cite{brinati2020detection,makridis2021designing,patelNatural2021,haimovich2020development,pennisiAn2021,jadhav_covid-view_2021}

User interfaces are a highly valuable way for users without significant computational experience to visualize the results of XAI approaches. These interfaces are frequently used to demonstrate model performance and to instill confidence in AI solutions. The following works \cite{brinati2020detection,makridis2021designing,patelNatural2021,haimovich2020development,pennisiAn2021,jadhav_covid-view_2021} all successfully leveraged XAI and user interfaces to deliver value to their clinician end users. Meanwhile, it enables the future collection of user feedback and clinical response.

Markridis et al. \cite{makridis2021designing} used XGBoost to predict the probability of patient mortality from a Veterans Affairs hospital dataset. In addition to feature importance derived directly from the XGBoost model, they also used SHAP to establish feature effects on prediction directionality. These insights were presented with a user interface for clinicians to examine individual patients and understand individual feature effects on final model prediction.

Brinati et al. \cite{brinati2020detection} used 13 clinical features (e.g. diagnostic blood lab values and patient gender) to compare seven machine learning models including random forest and conventional decision tree approaches using nested cross validation. Random forest was found to produce the best average accuracy across the external folds, and was also used to generate feature importance scores. The decision tree model, despite not performing as well, was also used to illustrate the contribution of features to the final classification label. The final workflow was made available to clinicians via a web-based user interface. 

Haimovich et al. \cite{haimovich2020development} established their outcome variable of patient respiratory compensation based on clinical insights and designed a web-based platform to be utilized in real-world clinical settings. They generated feature importance by comparing values obtained from multiple interpretable machine learning approaches. The ranked list of feature importance was also supplemented by SHAP to elucidate the direction of contribution of each feature across patients for clinical use.
\section{Evaluation of Explainable AI Methods}
\label{sec:evaluation}

\begin{figure}[]
    \centering
    \includegraphics[width=\linewidth]{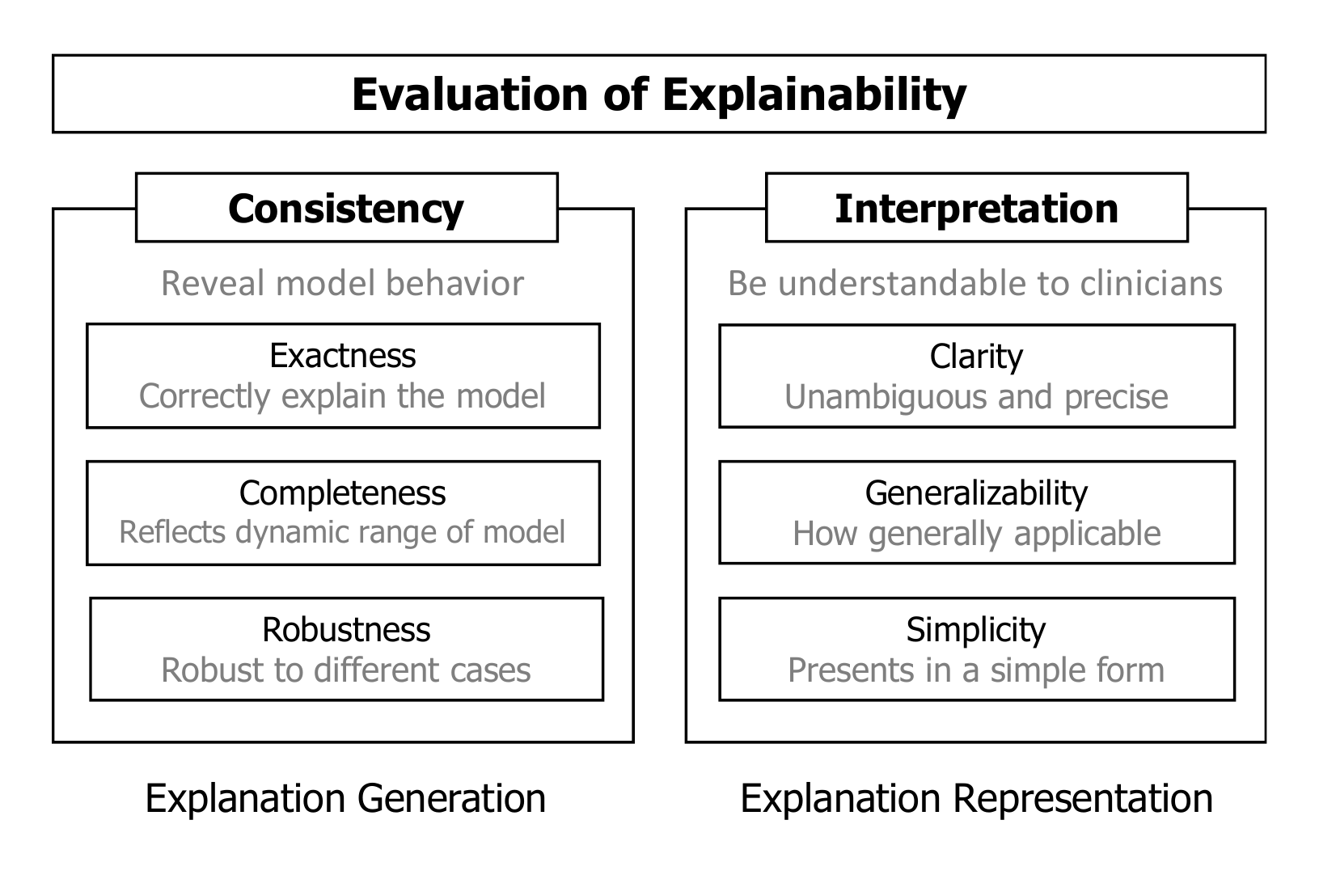}
    \vspace{-10mm}
    \caption{Evaluation of model explainability. To assess the generation of explanations and the revealing of model behaviors, it is critical to consider their correctness, completeness, and robustness. When evaluating explanation representations, it is critical to consider their clarity of presentation, their generalizability, and their simplicity of form.}
    \label{fig:evaluation}
\end{figure}

Qualitative visualization plays an important role in evaluating XAI methods. For biomedical applications, qualitative evaluation focuses on whether visualization could align with established knowledge. For instance, expert radiologists can assess how well the generated attention map identifies image regions of high diagnostic relevance \cite{tsiknakis2020interpretable}. A standard of evaluation is illustrated in Fig. \ref{fig:evaluation}. 

Although qualitative evaluation is important, quantitative evaluation of interpretation is still desirable. Quantitative evaluation can be obtained through either user study or automatic approaches. In the scenario of conducting user studies for feature scoring, target users (e.g. physicians for medical applications) perform certain tasks with, and without, the assistance of visual interpretation. For example, in using histopathology image for clinical diagnosis, clinicians will be asked to diagnose cases of original images versus images with visual interpretation. Then improvement in performance is measured with the assistance of visual interpretation. User studies could be considered as the most reliable approach for evaluating interpretability, if they are designed to resemble real application scenarios. However, conducting such user studies are expensive and time-consuming, especially for biomedical applications. 

An alternative approach is automatic evaluation, which serves as a proxy for user study without involving real users. Zeilar and Fergus first introduced the idea of the occlusion experiment \cite{zeiler2014visualizing}, in which portions of input images were systematically occluded by a grey square for monitoring the performance of deep learning models. Samek et al. \cite{samek2016evaluating} further formalized the occlusion experiments by introducing a procedure called “pixel flipping”, which destroys data points ordered by their feature importance scores and compares the decrease in classification metrics among multiple interpretation methods. A larger decrease in the metrics suggests a better interpretation method. Because occlusion experiments are model agnostic, they can be used as an objective measure for interpretation methods. On the other hand, the occlusion experiments can not serve as objective evaluation for perturbation-based feature scoring methods, such as Randomized Input Sampling for Explanation (RISE) \cite{petsiuk2018rise}, that perturb input directly to identify important features. 

Quantitative evaluation of data synthesis is still in its infancy. DeVries et al. \cite{devries2019evaluation} designed an evaluation metric, named Fréchet Joint Distance (FJD), for the quality of images generated by conditional GAN based on visual quality, intra-conditioning diversity, and conditional consistency. Assuming the joint distribution of hidden space and labels are Gaussian, they used FJD to compare the mean and variance between real and generated images. Recently, Yang et al. \cite{yang2019bim} created a ground-truth dataset consisting of mosaic natural images for interpretation methods and tried to unify the evaluation of both feature scoring and data synthesis methods. Their aforementioned methods are early in their developmental stage, even for natural images, and ways to adopt them into biomedical images and other biomedical data modalities remains an ongoing challenge. Besides, Lin et al. \cite{10.1145/3447548.3467213} proposed an adversarial attack to evaluate the robustness of interpretability in XAI methods by checking whether they can detect backdoor triggers present in the input. Researchers employed data poisoning to create trojaned models, generated saliency maps that will highlight the trigger to evaluates the saliency map output with three quantitative evaluation metrics (IoU, recovery rate, and recovering difference).
\section{Discussion}
\label{sec:discussion}

\begin{table*}[]
% \tiny
\centering
\caption{Summary of XAI Methods in Clinical Applications}
\label{tab:XAI_methods}
\resizebox{\textwidth}{!}{%
\begin{tabular}{|l|l|l|}
\hline
\textbf{Strategy} & \textbf{Category} & \textbf{Technique} \\ \hline
\multirow{2}{*}{Data Augmentation} & \multirow{2}{*}{Latent Space Interpretation} & Intrinsic latent space guidance \cite{Singh2021-qg,Waheed2020-fo,Loey2020-jb,Voynov2020-kz,rahmanAdversarial2021,palatnik_de_sousa_explainable_2021} \\ \cline{3-3} 
 &  & Post-hoc PCA-based \cite{Harkonen2020-ch} \\ \hline
\multirow{14}{*}{Outcome Prediction} & \multirow{3}{*}{Perturbation-based} & Feature occlusion and ablation \cite{zeiler2014visualizing,tang2019interpretable,gomes_features_2021,casiraghiExplainable2020,schirrmeister2017deep,hossain2019applying} \\ \cline{3-3} 
 &  & SHAP feature importance \cite{vstrumbelj2014explaining, lundberg2017unified, singh2020interpretation,lundberg2018explainable,zoabi2020covid,wu2020ai,ong_comparative_2021,torres2020pandemyc, vaid2020machine, foieni2020derivation, pan2020prognostic, tang2020prediction, zheng2021interpretable, makridis2021designing, haimovich2020development, abdulaal2020prognostic, cavallaro2021contrasting,cao2020impact,alves2021explaining} \\ \cline{3-3} 
 &  & Local interpretable model-agnostic explanations (LIME) \cite{ahsan2020study, ong_comparative_2021, ahsan_detection_2021, ye_explainable_2021,ribeiro2016should} \\ \cline{2-3} 
 & \multirow{2}{*}{Activation-based} & Activation maximization \cite{erhan2010understanding} \\ \cline{3-3} 
 &  & Class activation maps (CAM)\cite{zhou2016learning,han2020accurate} \\ \cline{2-3} 
 & \multirow{4}{*}{Gradient-based} & Gradient-based class score \cite{simonyan2013deep} \\ \cline{3-3} 
 &  & Deconvnet \cite{springenberg2014striving} \\ \cline{3-3} 
 &  & Layer-wise relevance propagation (LRP) \cite{bach2015pixel,karim2020deepcovidexplainer, bassi2021deep} \\ \cline{3-3} 
 &  & Gradient-based saliency analysis \cite{simonyan2013deep,shamout2021artificial,wu2021jcs, barbosa2021machine, qian2020m, singh2021interpretable,singh_screening_2021} \\ \cline{2-3} 
 & Mixed-based & Grad-CAM and Grad-CAM++ \cite{li2020using, wang2021psspnn, song2021deep, arias2020artificial, brunese2020explainable, panwar2020deep, alshazly2021explainable,selvaraju2017grad, chattopadhay2018grad,oh2020deep} \\ \cline{2-3} 
 & \multirow{4}{*}{Attention-based} & Self-attention mechanism \cite{vaswani2017attention,wang2018non,chen2019lesion,xu2019pulmonary,chikontwe_dual_2021,shi_covid-19_2021,shickel2019deepsofa} \\ \cline{3-3} 
 &  & Hierarchical attention mechanism \cite{dong2021explainable,zhang2018patient2vec} \\ \cline{3-3} 
 &  & RNN-based temporal attention mechanism \cite{choi2016retain, kwon2018retainvis, kaji2019attention, shickel2019deepsofa, yin2019domain, li2020marrying, chen2020interpretable} \\ \cline{3-3} 
 &  & Temporal convolution network \cite{lauritsen2020explainable} \\ \hline
\multirow{4}{*}{Unsupervised Clustering} & Guided Embedding & VAE-feature clustering \cite{song2013auto,lim2020deep,prasad2020variational} \\ \cline{2-3} 
 & \multirow{3}{*}{Feature Extraction} & Self-organizing feature map (SOFM) clustering \cite{king2020unsupervised} \\ \cline{3-3} 
 &  & Similarity calculation \cite{hegde2019similar,singh_these_2021} \\ \cline{3-3} 
 &  & Latent space interpretation \cite{yadav_lung-gans_2021} \\ \hline
\multirow{4}{*}{Image Segmentation} & Morphology-based & Maximum entropy threshold \cite{yao_csgbbnet_2021,nedumkunnel_explainable_2021,jadhav_covid-view_2021} \\ \cline{2-3} 
 & \multirow{3}{*}{Context-based} & Mult-scale attention \cite{wang_joint_2021}\\ \cline{3-3} 
 &  & Saliency Analysis \cite{wu2021jcs}\\ \cline{3-3} 
 &  & Network dissection \cite{natekar2020demystifiying} \\ \hline
\end{tabular}%
}
\end{table*}

Upon review of the existing works leveraging XAI to facilitate the interpretation of AI-based COVID-19 solutions to clinical challenges, we have identified key features present in papers which have made substantial impacts in the field. Table \ref{tab:XAI_methods} summarizes the XAI techniques used in COVID-19 related clinical applications in this review paper. Furthermore, we discuss these findings, and references to example implementations, as a table of important considerations during the process of AI-based experimental design Table. \ref{tab:checklist}.

\subsection{Checklist for AI-enabled Clinical Applications}
Using the framework values of performance, user trust, and user response, we noticed the need for incorporating clinical insights throughout the study design process. This includes understanding the factors influencing response variables in the real world, as illustrated in Haimovich et al. \cite{haimovich2020development} when they stated that ICU admission was not an ideal outcome variable due to site-specific and time-dependent patient admission requirements. Clinical input may also be obtained during and after model optimization via real-time expert feedback \cite{pennisi2021explainable} and during implementation via expert-facing user interfaces \cite{brinatiDetection2020,makridisDesigning2021,patelNatural2021,hou_explainable_2021}. In addition to web-based applications, visualizing sample clusters \cite{sanchez2020machine,arias2020artificial} and feature importance metrics \cite{oh2020deep,Zheng2019-bt,tang2020prediction,ramchandani2020deepcovidnet,pan2020prognostic,cao2020impact,dong2021explainable,cavallaro2021contrasting,casiraghiExplainable2020,abdulaal2020prognostic} can offer users without expertise in data analysis an option of understanding the decision-making process of otherwise obscure models. 

A very common approach to generating easily interpretable models is to optimize a decision tree approach that can describe the decision making process using available features \cite{brinati2020detection,alves2021explaining,torres2020pandemyc,vaid2020machine,foieni2020derivation,sanchez2020machine,casiraghiExplainable2020,pan2020prognostic,tang2020prediction,zheng2021interpretable,makridis2021designing,haimovich2020development,abdulaal2020prognostic,cavallaro2021contrasting}. This approach is also similar to commonly used clinical guidelines to generate fast and consistent metrics for patient triage and management \cite{jones2009sequential,barlow2007curb65}.
 
Validating feature importance ranking by using multiple methods, such as tree-based importance metrics and SHAPley values, can establish features lists which are consistent between approaches to prevent spurious rankings \cite{haimovich2020development,cavallaro2021contrasting,brinati2020detection,alves2021explaining,makridis2021designing,cao2020impact}. This may be especially important if the list is to be used for feature selection or simplified feature visualizations (e.g. only visualize odds ratios for most important features). 

Comparison of multiple competing models is often necessary to generate high-performance solutions. We noticed the widespread use of cross validation when authors sought to conduct these comparisons \cite{haimovich2020development,brinati2020detection,alves2021explaining,estiri2021individualized,abdulaal2020prognostic,pan2020prognostic}. Cross validation is easier to implement when models are quickly trained and tested, but this approach may also be used with more complex models to ensure robust comparisons.

In the quest for easily interpretable results, it is common to see accuracy being reported as a model performance metric. Although accuracy is understood by model developers and end-users alike, it should be avoided when significant data imbalance is present. Examples of works using appropriate performance metrics include \cite{estiri2021individualized,zheng2021interpretable,tang2020prediction,pan2020prognostic,abdulaal2020prognostic}. Common metrics include Area Under the Receiver Operating Curve (AUROC) and Matthews Correlation Coefficient (MCC), the later being appropriate even in imbalanced binary classification tasks \cite{chicco2020advantages}. 
 
 An often overlooked aspect of model development is the potential for adversarial attack within the final implementation context. Cyber attacks on hospital systems are depressingly common with a notable rise in frequency over time \cite{argaw2020cybersecurity}. As research tools make their way into the hospital it may become important to understand the vulnerability of models to potential future attacks. Therefore, we included this component to our checklist alongside a recent illustration of adversarial testing approaches \cite{rahmanAdversarial2021}.

% Challenges
\subsection{Challenges and Solutions}
\begin{table*}[]
\centering
\caption{Checklist for AI-enabled Clinical Decision Support Systems}
\label{tab:checklist}
\begin{tabular}{@{}llll@{}}
\cmidrule(r){1-1} \cmidrule(l){3-4}
% \multicolumn{1}{c}{\textit{\textbf{Insights}}} & \multicolumn{1}{c}{} & \multicolumn{2}{c}{\textit{\textbf{Challenges}}} \\ \cmidrule(r){1-1} \cmidrule(l){3-4} 
\multicolumn{1}{c}{\textbf{Suggestions}} &  & \multicolumn{1}{c|}{\textbf{Problems}} & \multicolumn{1}{c}{\textbf{Solutions}} \\ \cmidrule(r){1-1} \cmidrule(l){3-4} 
Incorporate clinical insights \cite{haimovich2020development}&  & Small sample size & Data imputation \cite{barbosa2021machine,gomes_features_2021} \\
Interactive user interface \cite{brinati2020detection,makridis2021designing,patelNatural2021,haimovich2020development,pennisiAn2021,jadhav_covid-view_2021}&  & Bad data quality & Artifact correction \cite{li2020using} \\
Clinical feedback \cite{pennisiAn2021}&  & Imbalanced classes & Data augmentation \cite{wang2021psspnn,singh_screening_2021}\\
Visualization of feature importance \cite{oh2020deep,tang2020prediction,pan2020prognostic,cao2020impact,dong2021explainable,cavallaro2021contrasting,abdulaal2020prognostic} &  & Complex disease phenotype & Multi-modality data \cite{shamout2021artificial}\\
Clustering analysis \cite{sanchez2020machine,arias2020artificial,yadav_lung-gans_2021} &  & Data heterogenity & Data normalization \cite{barbosa2021machine}\\
Decision tree \cite{brinati2020detection,alves2021explaining,torres2020pandemyc,vaid2020machine,foieni2020derivation,sanchez2020machine,casiraghiExplainable2020,pan2020prognostic,tang2020prediction,zheng2021interpretable,makridis2021designing,haimovich2020development,abdulaal2020prognostic,cavallaro2021contrasting} &  & Lack of expert annotation & Weakly supervised learning \cite{han2020accurate,oh2020deep,yadav_lung-gans_2021}\\
Use multiple feature importance approaches \cite{cavallaro2021contrasting,brinati2020detection,alves2021explaining,makino2020differences,ong_comparative_2021,chikontwe_dual_2021} &  & Unkown sources of signal & Key feature extraction \cite{panwar2020deep,singh2020interpretation,wu2021jcs}\\
Cross validation when comparing models \cite{haimovich2020development,brinati2020detection,alves2021explaining,estiri2021individualized,abdulaal2020prognostic,pan2020prognostic,ahsan_detection_2021}&  & Explanations unclear & Pre-processing changes \cite{arias2020artificial,brunese2020explainable}\\
Use appropriate and robust performance metrics (AUROC, MCC) \cite{estiri2021individualized,zheng2021interpretable}&  & Data leakage & Patient-level split \cite{song2021deep,qian2020m,ahsan2020study,wang_joint_2021}\\
Adversarial example testing \cite{rahmanAdversarial2021,palatnik_de_sousa_explainable_2021}&  & Training is inefficient & Transfer learning \cite{bassi2021deep,yao_csgbbnet_2021,singh_screening_2021} \\ \cmidrule(r){1-1} \cmidrule(l){3-4} 
\end{tabular}
\end{table*}

\begin{figure*}[]
    \centering
    \includegraphics[width=\linewidth]{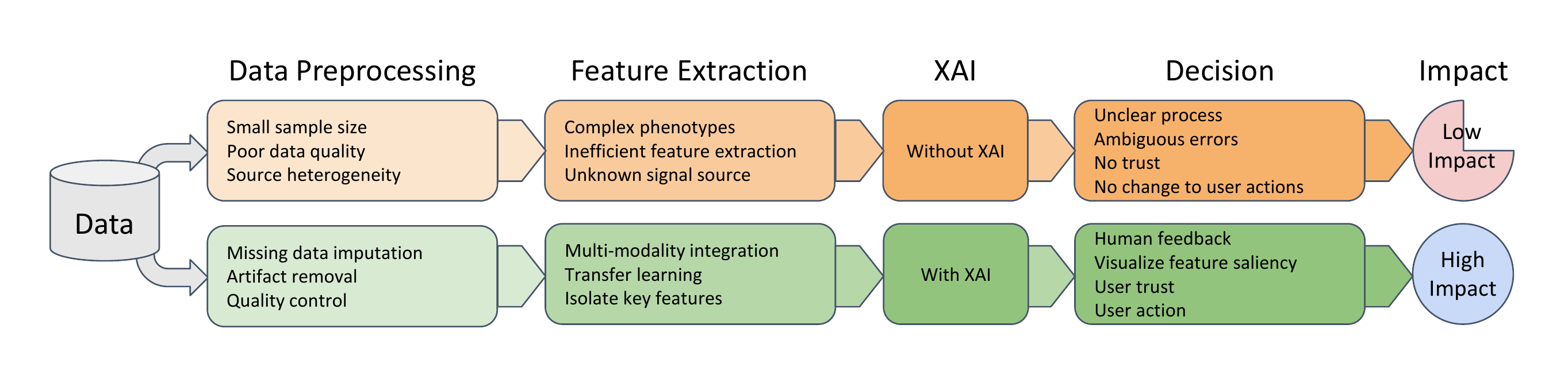}
    \vspace{-8mm}
    \caption{Summary of insights gained for designing an AI development workflow. We provide a checklist of considerations to make early in the experimental design process in order to avoid common problems. Additionally, we provide a list of common issues encountered when working with clinical data and discuss several common solutions that may assist the reader when working with these data.}
    \label{fig:workflow}
\end{figure*}

With any clinical informatics work there will be challenges. Often these will arise due to issues with the dataset being used, especially if it was derived from real-world data. After our review of the literature we summarized common challenges and potential solutions, including example works which successfully solve the problem Table. \ref{tab:checklist}. 

Early in the pandemic, there was a scarcity of reliable data available to the general scientific community. This resulted in a significant need for data imputation in order to fill in missing values to maximize the utility of existing data \cite{barbosa2021machine}.  Poor data quality also affected model performance and artifact correction techniques were implemented \cite{li2020using}.

Imbalanced classes were frequently found within COVID-19 datasets due to the accessibility of normal samples relative to COVID-19 positive. Data augmentation was found to alleviate this problem in some cases by generating additional samples of the underrepresented class \cite{wang2021psspnn}.

Lack of expert annotation of key regions of pathology in imaging data created the need for weakly supervised learning models capable of generalizing small ground truth datasets \cite{han2020accurate,oh2020deep}. Without expert insight, it was often necessary to identify features capable of differentiating between similar phenotypes (e.g. bacterial versus viral pneumonia). This problems was frequently solved via key feature extraction \cite{han2020accurate,li2020using,song2021deep,panwar2020deep,oh2020deep,qian2020m,singh2021interpretable,wu2021jcs}. In the case of complex disease phenotypes, multi-modality data were integrated to leverage data obtained from consistent or complementary sources \cite{shamout2021artificial}. 
Ensuring model generalizability requires robust external validation. Data leakage occurs when testing/external dataset information is used during the model design or optimization process. The likelihood of this occurring can be reduced by isolating the test dataset during hyperparameter selection and model training. Special care must be taken to avoid including data derived from the same patient in both the test and training datasets. There exists significant within-patient correlation of features, even across samples. This data leakage may allow models to learn patient-specific patterns which are not generalizable to other patients, resulting in poor performance in the real-world \cite{li2020using,song2021deep,qian2020m,ahsan2020study}. 

XAI may lead to unclear results, either due to inconsistent feature importance ranking or nonspecific image highlighting. In these cases, it is often a good idea to re-establish the quality of the preprocessing pipeline \cite{arias2020artificial, brunese2020explainable}.

When training is inefficient, transfer learning may be used to take advantage of prior parameter optimization on similar problems \cite{alshazly2021explainable,bassi2021deep}.

Ultimately, we designed this checklist to help both academic researchers in general and clinical data scientists specifically. We summarize the integration of XAI in both settings, along with its benefits in Figure \ref{fig:workflow}.

% Trends
\subsection{Evolution of XAI Methods}
XAI techniques have developed quickly in recent years to meet the evolving needs of AI researchers and the end-users of their models. Although it is easy to fall into the trap of believing that more recent models are objectively better than their more classic counterparts, it is important to understand that each model was designed to improve our understanding of different facets of AI-solutions. For example, in the task of data augmentation, it is common to see K-Nearest Neighbors interpolation and other classic approaches used instead of more complex modern solutions. This is in part because the more classic approach has been around for longer and its pitfalls have been well established. The modern approach may result in data bias which may be difficult to understand due to the lack of real-world examples of their successes and failures. The trend for data augmentation has been to increase the number of considered factors and complexity of data transformations to better model the underlying data distribution of samples. 

\begin{table*}[]
\centering
\caption{Summary of XAI Tools, Libraries, and Packages for Clinical Applications}
\label{tab:tools}
\begin{tabular}{@{}c|l@{}}
\toprule
\textbf{Tools/Packages} & \multicolumn{1}{c}{\textbf{Descriptions}}                                                                    \\ \midrule
SHAP                    & A game theoretic approach to explain the output of any ML model (https://github.com/slundberg/shap)          \\ \midrule
ELI5                 & A Python library for debugging/inspecting ML classifiers and explaining their predictions (https://github.com/TeamHG-Memex/eli5)      \\ \midrule
Skater                  & A Python library for model agnostic interpretation with unified framework (https://github.com/oracle/Skater)   \\ \midrule
alibi                   & Algorithms for monitoring and explaining ML models (https://github.com/SeldonIO/alibi)                       \\ \midrule
InterpretML             & An open-source package that incorporates state-of-the-art ML interpretability techniques (https://github.com/interpretml/interpret)                  \\ \midrule
Explanation Explorer & A user interface to interpret ML models and explore similarly explained data items (https://github.com/nyuvis/explanation$\_$explorer) \\ \midrule
AIX360                  & An open-source library to support interpretability of data and ML models (https://github.com/Trusted-AI/AIX360) \\ \midrule
yellowbrik              & An extension of scikit-learn library for visualization of ML models (https://www.scikit-yb.org/en/latest/)   \\ \midrule
tf-explain              & A Tensorflow 2.0 library for deep learning model interpretability (https://github.com/sicara/tf-explain)     \\ \bottomrule
\end{tabular}
\end{table*}

Clinical decision support is a very common setting to find AI-solutions in need of explanation. For the task of disease diagnosis, the trend has been to generate input importance visualizations which can be used across a wide range of common deep learning models. This is in contrast to early XAI approaches which relied on model-specific solutions to improve interpretability. XAI in risk prediction for clinical decision support has trended towards generating sample-specific explanations. These may provide the end user with a custom answer to the question of “why did this sample get the score that it did?”. This is especially useful in the clinical setting where precision medicine is becoming the standard, and patient-specific explanations for risk scores are vital. Additionally, we summarized in Table \ref{tab:tools} a list of the most popular XAI tools, libraries, and packages that can be used directly in clinical applications. 

Unsupervised clustering has benefited from the use of deep learning model latent spaces for their generation of sample similarities. This shift from the classical input feature distance approaches allows custom transformations to optimize the space within which similarity is measured. This can result in better disentanglement of samples \cite{kang_quantitative_2021}.  

Image segmentation approaches have increased in complexity in recent years due to models such as U-Net and its variants. Models have gone from highly interpretable (if using simple color thresholds) to involving many nonlinear transformations to producing the final segmentations. XAI approaches for image segmentation are still not commonly used, but recent techniques have leveraged the model activation maps produced by deep layers to identify significant associations with final deep learning model output \cite{Bau2020-rx}. 
XAI approaches will continue to adapt as models continue to become better optimized for different tasks. XAI will likely cover a much wider range of approaches to meet the needs of end-users and regulatory agencies. 

In future work, with the decrease of COVID-19 incidence and increase of vaccine supply, risk stratification will become vital to determine optimal treatment plan. We also hope our focus of XAI within the ongoing COVID-19 pandemic may increase the relevance of our insights to future disease outbreaks. The framework we provided can be used across common AI-tasks and may improve the clinical implementation of these solutions, especially in the early stages of infection.

\section{Conclusion}
\label{sec:conclusion}

The recent confluence of large-scale public healthcare datasets combined with the rapid increase of computing capacity, and the popularity of open-source code has resulted in a noteworthy increase in AI-based clinical decision support systems.
This trend has increased the need for understanding the criteria which make AI solutions successful in practice. 

In this work, we reviewed XAI approaches used to solve challenges which arose during the COVID-19 pandemic to identify trends and insights useful for meeting future challenges. 
First, we provided a general overview of common XAI techniques used during the pandemic and gave specific examples to demonstrate the value that these techniques provide. 
We then illustrated classical, recent, and potential future trends in XAI approaches to clarify the evolving goals of these tools. 
Evaluation approaches were also discussed to provide the reader with an understanding of qualitative and quantitative methods used to check the performance of XAI results.
After covering the different aspects of implementing XAI, we summarized the insights that we have gained for the design of an AI development workflow. We provided a checklist of suggestions to consider early during the experimental design process to avoid prevalent issues. We also provided a list of common problems seen when working with clinical data and discuss some common solutions which may aid the reader during their work with these data.
Finally, we discussed the potential challenges and future directions of XAI in COVID-19 and biomedical applications, with an ideal workflow meeting the requirements of performance, trustworthiness, and impact on user response.

Clinical informatics is generally risk-averse creating the need for AI developers in the field to understand how AI-based decisions are reached. This understanding would provide two benefits: 
(i) increasing confidence that a deep learning model is unbiased and relies on relevant features to accomplish desired tasks and 
(ii) detecting biases or discovering new knowledge if the generated explanations do not fit with established science. 
Ultimately, we hope that the implementation of XAI techniques will accelerate the translation of data-driven analytic solutions to improve the quality of patient care.

\section*{Acknowledgment}
The authors would like to thank the Emory Science Librarian, Ms. Kristan Majors, for her support and guidance on search optimization for the PRISMA chart. We would like to thank Dr. Siva Bhavani from Emory University for his insights on leveraging artificial intelligence in clinical practice. We would like to thank Mr. Benoit Marteau from Bio-MIBLab for his help on reviewing the manuscript. This research was supported by a Wallace H. Coulter Distinguished Faculty Fellowship (M. D. Wang), a Petit Institute Faculty Fellowship (M. D. Wang), and by Microsoft Research.

\bibliographystyle{IEEEtran}
\bibliography{IEEEabrv,mybib}

% Generated by IEEEtran.bst, version: 1.14 (2015/08/26)
\begin{thebibliography}{100}
\providecommand{\url}[1]{#1}
\csname url@samestyle\endcsname
\providecommand{\newblock}{\relax}
\providecommand{\bibinfo}[2]{#2}
\providecommand{\BIBentrySTDinterwordspacing}{\spaceskip=0pt\relax}
\providecommand{\BIBentryALTinterwordstretchfactor}{4}
\providecommand{\BIBentryALTinterwordspacing}{\spaceskip=\fontdimen2\font plus
\BIBentryALTinterwordstretchfactor\fontdimen3\font minus
  \fontdimen4\font\relax}
\providecommand{\BIBforeignlanguage}[2]{{%
\expandafter\ifx\csname l@#1\endcsname\relax
\typeout{** WARNING: IEEEtran.bst: No hyphenation pattern has been}%
\typeout{** loaded for the language `#1'. Using the pattern for}%
\typeout{** the default language instead.}%
\else
\language=\csname l@#1\endcsname
\fi
#2}}
\providecommand{\BIBdecl}{\relax}
\BIBdecl

\bibitem{hu2020characteristics}
B.~Hu, H.~Guo, P.~Zhou, and Z.-L. Shi, ``Characteristics of sars-cov-2 and
  covid-19,'' \emph{Nature Reviews Microbiology}, pp. 1--14, 2020.

\bibitem{rubin2020role}
G.~D. Rubin, C.~J. Ryerson, L.~B. Haramati, N.~Sverzellati, J.~P. Kanne,
  S.~Raoof, N.~W. Schluger, A.~Volpi, J.-J. Yim, I.~B. Martin \emph{et~al.},
  ``The role of chest imaging in patient management during the covid-19
  pandemic: a multinational consensus statement from the fleischner society,''
  \emph{Chest}, 2020.

\bibitem{soltan2020rapid}
A.~A. Soltan, S.~Kouchaki, T.~Zhu, D.~Kiyasseh, T.~Taylor, Z.~B. Hussain,
  T.~Peto, A.~J. Brent, D.~W. Eyre, and D.~A. Clifton, ``Rapid triage for
  covid-19 using routine clinical data for patients attending hospital:
  development and prospective validation of an artificial intelligence
  screening test,'' \emph{The Lancet Digital Health}, 2020.

\bibitem{yan2020interpretable}
L.~Yan, H.-T. Zhang, J.~Goncalves, Y.~Xiao, M.~Wang, Y.~Guo, C.~Sun, X.~Tang,
  L.~Jing, M.~Zhang \emph{et~al.}, ``An interpretable mortality prediction
  model for covid-19 patients,'' \emph{Nature Machine Intelligence}, pp. 1--6,
  2020.

\bibitem{Heaven2021-ef}
W.~D. Heaven, ``Hundreds of {AI} tools have been built to catch covid. none of
  them helped,'' \emph{MIT Technology Review}, Jul. 2021.

\bibitem{zhang2020clinically}
K.~Zhang, X.~Liu, J.~Shen, Z.~Li, Y.~Sang, X.~Wu, Y.~Zha, W.~Liang, C.~Wang,
  K.~Wang \emph{et~al.}, ``Clinically applicable ai system for accurate
  diagnosis, quantitative measurements, and prognosis of covid-19 pneumonia
  using computed tomography,'' \emph{Cell}, 2020.

\bibitem{li2020using}
L.~Li, L.~Qin, Z.~Xu, Y.~Yin, X.~Wang, B.~Kong, J.~Bai, Y.~Lu, Z.~Fang, Q.~Song
  \emph{et~al.}, ``Using artificial intelligence to detect covid-19 and
  community-acquired pneumonia based on pulmonary ct: evaluation of the
  diagnostic accuracy,'' \emph{Radiology}, vol. 296, no.~2, 2020.

\bibitem{minaee2020deep}
S.~Minaee, R.~Kafieh, M.~Sonka, S.~Yazdani, and G.~J. Soufi, ``Deep-covid:
  Predicting covid-19 from chest x-ray images using deep transfer learning,''
  \emph{Medical image analysis}, vol.~65, p. 101794, 2020.

\bibitem{Rahman2021-ax}
A.~Rahman, M.~S. Hossain, N.~A. Alrajeh, and F.~Alsolami, ``Adversarial
  {Examples---Security} threats to {COVID-19} deep learning systems in medical
  {IoT} devices,'' \emph{IEEE Internet of Things Journal}, vol.~8, no.~12, pp.
  9603--9610, Jun. 2021.

\bibitem{su2019one}
J.~Su, D.~V. Vargas, and K.~Sakurai, ``One pixel attack for fooling deep neural
  networks,'' \emph{IEEE Transactions on Evolutionary Computation}, vol.~23,
  no.~5, pp. 828--841, 2019.

\bibitem{eykholt2018robust}
K.~Eykholt, I.~Evtimov, E.~Fernandes, B.~Li, A.~Rahmati, C.~Xiao, A.~Prakash,
  T.~Kohno, and D.~Song, ``Robust physical-world attacks on deep learning
  visual classification,'' in \emph{Proceedings of the IEEE Conference on
  Computer Vision and Pattern Recognition}, 2018, pp. 1625--1634.

\bibitem{alvarez2018towards}
D.~Alvarez-Melis and T.~S. Jaakkola, ``Towards robust interpretability with
  self-explaining neural networks,'' in \emph{Proceedings of the 32nd
  International Conference on Neural Information Processing Systems}, 2018, pp.
  7786--7795.

\bibitem{Cruz_Rivera2020-al}
S.~Cruz~Rivera, X.~Liu, A.-W. Chan, A.~K. Denniston, M.~J. Calvert, and
  {SPIRIT-AI and CONSORT-AI Working Group},
  ``\BIBforeignlanguage{en}{Guidelines for clinical trial protocols for
  interventions involving artificial intelligence: the {SPIRIT-AI}
  extension},'' \emph{\BIBforeignlanguage{en}{Lancet Digit Health}}, vol.~2,
  no.~10, pp. e549--e560, Oct. 2020.

\bibitem{Liu2020-jm}
X.~Liu, S.~Cruz~Rivera, D.~Moher, M.~J. Calvert, A.~K. Denniston, and
  {SPIRIT-AI and CONSORT-AI Working Group}, ``\BIBforeignlanguage{en}{Reporting
  guidelines for clinical trial reports for interventions involving artificial
  intelligence: the {CONSORT-AI} extension},''
  \emph{\BIBforeignlanguage{en}{Lancet Digit Health}}, vol.~2, no.~10, pp.
  e537--e548, Oct. 2020.

\bibitem{makino2020differences}
T.~Makino, S.~Jastrzebski, W.~Oleszkiewicz, C.~Chacko, R.~Ehrenpreis,
  N.~Samreen, C.~Chhor, E.~Kim, J.~Lee, K.~Pysarenko \emph{et~al.},
  ``Differences between human and machine perception in medical diagnosis,''
  \emph{arXiv preprint arXiv:2011.14036}, 2020.

\bibitem{moher2009preferred}
D.~Moher, A.~Liberati, J.~Tetzlaff, D.~G. Altman, P.~Group \emph{et~al.},
  ``Preferred reporting items for systematic reviews and meta-analyses: the
  prisma statement,'' \emph{PLoS medicine}, vol.~6, no.~7, p. e1000097, 2009.

\bibitem{Singh2021-qg}
R.~K. Singh, R.~Pandey, and R.~N. Babu,
  ``\BIBforeignlanguage{en}{{{COVIDScreen}}: explainable deep learning
  framework for differential diagnosis of {{COVID}-19} using chest {X}-rays},''
  \emph{\BIBforeignlanguage{en}{Neural Comput. Appl.}}, pp. 1--22, Jan. 2021.

\bibitem{Waheed2020-fo}
A.~Waheed, M.~Goyal, D.~Gupta, A.~Khanna, F.~Al-Turjman, and P.~R. Pinheiro,
  ``{CovidGAN}: Data augmentation using auxiliary classifier {GAN} for improved
  covid-19 detection,'' \emph{IEEE Access}, vol.~8, pp. 91\,916--91\,923, 2020.

\bibitem{Loey2020-jb}
M.~Loey, G.~Manogaran, and N.~E.~M. Khalifa, ``\BIBforeignlanguage{en}{A deep
  transfer learning model with classical data augmentation and {CGAN} to detect
  {COVID-19} from chest {CT} radiography digital images},''
  \emph{\BIBforeignlanguage{en}{Neural Comput. Appl.}}, pp. 1--13, Oct. 2020.

\bibitem{Voynov2020-kz}
A.~Voynov and A.~Babenko, ``Unsupervised discovery of interpretable directions
  in the gan latent space,'' in \emph{International Conference on Machine
  Learning}.\hskip 1em plus 0.5em minus 0.4em\relax PMLR, 2020, pp. 9786--9796.

\bibitem{noauthor_undated-zz}
L.~Deng, ``The mnist database of handwritten digit images for machine learning
  research,'' \emph{IEEE Signal Processing Magazine}, vol.~29, no.~6, pp.
  141--142, 2012.

\bibitem{Jin_undated-yd}
Y.~Jin, J.~Zhang, M.~Li, Y.~Tian, and H.~Zhu, ``Towards the high-quality anime
  characters generation with generative adversarial networks,'' in
  \emph{Proceedings of the Machine Learning for Creativity and Design Workshop
  at NIPS}, 2017.

\bibitem{Liu_undated-yv}
Z.~Liu, P.~Luo, X.~Wang, and X.~Tang, ``Deep learning face attributes in the
  wild,'' in \emph{Proceedings of the IEEE international conference on computer
  vision}, 2015, pp. 3730--3738.

\bibitem{Brock2019-nt}
A.~Brock, J.~Donahue, and K.~Simonyan, ``Large scale {GAN} training for high
  fidelity natural image synthesis,'' in \emph{International Conference on
  Learning Representations}, 2019.

\bibitem{Harkonen2020-ch}
E.~H{\"a}rk{\"o}nen, A.~Hertzmann, J.~Lehtinen, and S.~Paris, ``Ganspace:
  Discovering interpretable gan controls,'' \emph{Advances in Neural
  Information Processing Systems}, vol.~33, 2020.

\bibitem{rahmanAdversarial2021}
A.~Rahman, M.~S. Hossain, N.~A. Alrajeh, and F.~Alsolami, ``Adversarial
  examples--security threats to covid-19 deep learning systems in medical iot
  devices,'' \emph{IEEE Internet of Things Journal}, 2020.

\bibitem{palatnik_de_sousa_explainable_2021}
I.~Palatnik~de Sousa, M.~M. Vellasco, and E.~Costa~da Silva, ``Explainable
  artificial intelligence for bias detection in covid ct-scan classifiers,''
  \emph{Sensors}, vol.~21, no.~16, p. 5657, 2021.

\bibitem{zeiler2014visualizing}
M.~D. Zeiler and R.~Fergus, ``Visualizing and understanding convolutional
  networks,'' in \emph{European conference on computer vision}.\hskip 1em plus
  0.5em minus 0.4em\relax Springer, 2014, pp. 818--833.

\bibitem{tang2019interpretable}
Z.~Tang, K.~V. Chuang, C.~DeCarli, L.-W. Jin, L.~Beckett, M.~J. Keiser, and
  B.~N. Dugger, ``Interpretable classification of alzheimer’s disease
  pathologies with a convolutional neural network pipeline,'' \emph{Nature
  communications}, vol.~10, no.~1, pp. 1--14, 2019.

\bibitem{gomes_features_2021}
D.~P. Gomes, A.~Ulhaq, M.~Paul, M.~J. Horry, S.~Chakraborty, M.~Saha,
  T.~Debnath, and D.~M. Rahaman, ``Features of icu admission in x-ray images of
  covid-19 patients,'' in \emph{2021 IEEE International Conference on Image
  Processing}, 2021.

\bibitem{casiraghiExplainable2020}
E.~Casiraghi, D.~Malchiodi, G.~Trucco, M.~Frasca, L.~Cappelletti, T.~Fontana,
  A.~A. Esposito, E.~Avola, A.~Jachetti, J.~Reese \emph{et~al.}, ``Explainable
  machine learning for early assessment of covid-19 risk prediction in
  emergency departments,'' \emph{IEEE Access}, vol.~8, pp. 196\,299--196\,325,
  2020.

\bibitem{vstrumbelj2014explaining}
E.~{\v{S}}trumbelj and I.~Kononenko, ``Explaining prediction models and
  individual predictions with feature contributions,'' \emph{Knowledge and
  information systems}, vol.~41, no.~3, pp. 647--665, 2014.

\bibitem{lundberg2017unified}
S.~M. Lundberg and S.-I. Lee, ``A unified approach to interpreting model
  predictions,'' in \emph{Advances in neural information processing systems},
  2017, pp. 4765--4774.

\bibitem{singh2020interpretation}
A.~Singh, A.~R. Mohammed, J.~Zelek, and V.~Lakshminarayanan, ``Interpretation
  of deep learning using attributions: application to ophthalmic diagnosis,''
  in \emph{Applications of Machine Learning 2020}, vol. 11511.\hskip 1em plus
  0.5em minus 0.4em\relax International Society for Optics and Photonics, 2020,
  p. 115110A.

\bibitem{lundberg2018explainable}
S.~M. Lundberg, B.~Nair, M.~S. Vavilala, M.~Horibe, M.~J. Eisses, T.~Adams,
  D.~E. Liston, D.~K.-W. Low, S.-F. Newman, J.~Kim \emph{et~al.}, ``Explainable
  machine-learning predictions for the prevention of hypoxaemia during
  surgery,'' \emph{Nature biomedical engineering}, vol.~2, no.~10, pp.
  749--760, 2018.

\bibitem{zoabi2020covid}
Y.~Zoabi, S.~Deri-Rozov, and N.~Shomron, ``Machine learning-based prediction of
  covid-19 diagnosis based on symptoms,'' \emph{npj digital medicine}, vol.~4,
  no.~1, pp. 1--5, 2021.

\bibitem{wu2020ai}
H.~Wu, W.~Ruan, J.~Wang, D.~Zheng, S.~Li, J.~Chen, K.~Li, X.~Chai, and
  S.~Helal, ``Is ai model interpretable to combat with covid? an empirical
  study on severity prediction task,'' \emph{arXiv preprint arXiv:2010.02006},
  2020.

\bibitem{ong_comparative_2021}
J.~H. Ong, K.~M. Goh, and L.~L. Lim, ``Comparative analysis of explainable
  artificial intelligence for covid-19 diagnosis on cxr image,'' in \emph{2021
  IEEE International Conference on Signal and Image Processing Applications
  (ICSIPA)}.\hskip 1em plus 0.5em minus 0.4em\relax IEEE, 2021, pp. 185--190.

\bibitem{ribeiro2016should}
M.~T. Ribeiro, S.~Singh, and C.~Guestrin, ``" why should i trust you?"
  explaining the predictions of any classifier,'' in \emph{Proceedings of the
  22nd ACM SIGKDD international conference on knowledge discovery and data
  mining}, 2016, pp. 1135--1144.

\bibitem{ahsan2020study}
M.~M. Ahsan, K.~D. Gupta, M.~M. Islam, S.~Sen, M.~Rahman, M.~S. Hossain
  \emph{et~al.}, ``Study of different deep learning approach with explainable
  ai for screening patients with covid-19 symptoms: Using ct scan and chest
  x-ray image dataset,'' \emph{arXiv preprint arXiv:2007.12525}, 2020.

\bibitem{ahsan_detection_2021}
M.~M. Ahsan, R.~Nazim, Z.~Siddique, and P.~Huebner, ``Detection of covid-19
  patients from ct scan and chest x-ray data using modified mobilenetv2 and
  lime,'' in \emph{Healthcare}, vol.~9, no.~9.\hskip 1em plus 0.5em minus
  0.4em\relax Multidisciplinary Digital Publishing Institute, 2021, p. 1099.

\bibitem{ye_explainable_2021}
Q.~Ye, J.~Xia, and G.~Yang, ``Explainable {AI} for {COVID}-19 {CT} classifiers:
  An initial comparison study,'' in \emph{2021 {IEEE} 34th International
  Symposium on Computer-Based Medical Systems ({CBMS})}, 2021, pp. 521--526,
  {ISSN}: 2372-9198.

\bibitem{erhan2010understanding}
D.~Erhan, A.~Courville, and Y.~Bengio, ``Understanding representations learned
  in deep architectures,'' \emph{Department dInformatique et Recherche
  Operationnelle, University of Montreal, QC, Canada, Tech. Rep}, vol. 1355,
  p.~1, 2010.

\bibitem{zhou2016learning}
B.~Zhou, A.~Khosla, A.~Lapedriza, A.~Oliva, and A.~Torralba, ``Learning deep
  features for discriminative localization,'' in \emph{Proceedings of the IEEE
  conference on computer vision and pattern recognition}, 2016, pp. 2921--2929.

\bibitem{han2020accurate}
Z.~Han, B.~Wei, Y.~Hong, T.~Li, J.~Cong, X.~Zhu, H.~Wei, and W.~Zhang,
  ``Accurate screening of covid-19 using attention-based deep 3d multiple
  instance learning,'' \emph{IEEE transactions on medical imaging}, vol.~39,
  no.~8, pp. 2584--2594, 2020.

\bibitem{springenberg2014striving}
J.~T. Springenberg, A.~Dosovitskiy, T.~Brox, and M.~Riedmiller, ``Striving for
  simplicity: The all convolutional net,'' \emph{arXiv preprint
  arXiv:1412.6806}, 2014.

\bibitem{simonyan2013deep}
K.~Simonyan, A.~Vedaldi, and A.~Zisserman, ``Deep inside convolutional
  networks: Visualising image classification models and saliency maps,'' in
  \emph{In Workshop at International Conference on Learning
  Representations}.\hskip 1em plus 0.5em minus 0.4em\relax Citeseer, 2014.

\bibitem{bach2015pixel}
S.~Bach, A.~Binder, G.~Montavon, F.~Klauschen, K.-R. M{\"u}ller, and W.~Samek,
  ``On pixel-wise explanations for non-linear classifier decisions by
  layer-wise relevance propagation,'' \emph{PloS one}, vol.~10, no.~7, p.
  e0130140, 2015.

\bibitem{karim2020deepcovidexplainer}
M.~R. Karim, T.~D{\"o}hmen, M.~Cochez, O.~Beyan, D.~Rebholz-Schuhmann, and
  S.~Decker, ``Deepcovidexplainer: Explainable covid-19 diagnosis from chest
  x-ray images,'' \emph{2020 IEEE International Conference on Bioinformatics
  and Biomedicine (BIBM)}, pp. 1034--1037, 2020.

\bibitem{bassi2021deep}
P.~R. Bassi and R.~Attux, ``A deep convolutional neural network for covid-19
  detection using chest x-rays,'' \emph{Research on Biomedical Engineering},
  pp. 1--10, 2021.

\bibitem{shamout2021artificial}
F.~E. Shamout, Y.~Shen, N.~Wu, A.~Kaku, J.~Park, T.~Makino, S.~Jastrzebski,
  J.~Witowski, D.~Wang, B.~Zhang \emph{et~al.}, ``An artificial intelligence
  system for predicting the deterioration of covid-19 patients in the emergency
  department,'' \emph{NPJ digital medicine}, vol.~4, no.~1, pp. 1--11, 2021.

\bibitem{wu2021jcs}
Y.~H. {Wu}, S.~H. {Gao}, J.~{Mei}, J.~{Xu}, D.~P. {Fan}, R.~G. {Zhang}, and
  M.~M. {Cheng}, ``Jcs: An explainable covid-19 diagnosis system by joint
  classification and segmentation,'' \emph{IEEE Transactions on Image
  Processing}, vol.~30, pp. 3113--3126, 2021.

\bibitem{barbosa2021machine}
E.~J.~M. Barbosa, B.~Georgescu, S.~Chaganti, G.~B. Aleman, J.~B. Cabrero,
  G.~Chabin, T.~Flohr, P.~Grenier, S.~Grbic, N.~Gupta \emph{et~al.}, ``Machine
  learning automatically detects covid-19 using chest cts in a large
  multicenter cohort,'' \emph{European radiology}, pp. 1--11, 2021.

\bibitem{qian2020m}
X.~Qian, H.~Fu, W.~Shi, T.~Chen, Y.~Fu, F.~Shan, and X.~Xue, ``M
  \textsuperscript{3} lung-sys: A deep learning system for multi-class lung
  pneumonia screening from ct imaging,'' \emph{IEEE journal of biomedical and
  health informatics}, vol.~24, no.~12, pp. 3539--3550, 2020.

\bibitem{singh2021interpretable}
G.~Singh and K.-C. Yow, ``An interpretable deep learning model for covid-19
  detection with chest x-ray images,'' \emph{IEEE Access}, 2021.

\bibitem{singh_screening_2021}
D.~Singh, V.~Kumar, M.~Kaur, M.~Y. Jabarulla, and H.-N. Lee, ``Screening of
  covid-19 suspected subjects using multi-crossover genetic algorithm based
  dense convolutional neural network,'' \emph{IEEE Access}, 2021.

\bibitem{selvaraju2017grad}
R.~R. Selvaraju, M.~Cogswell, A.~Das, R.~Vedantam, D.~Parikh, and D.~Batra,
  ``Grad-cam: Visual explanations from deep networks via gradient-based
  localization,'' in \emph{Proceedings of the IEEE international conference on
  computer vision}, 2017, pp. 618--626.

\bibitem{chattopadhay2018grad}
A.~Chattopadhay, A.~Sarkar, P.~Howlader, and V.~N. Balasubramanian,
  ``Grad-cam++: Generalized gradient-based visual explanations for deep
  convolutional networks,'' in \emph{2018 IEEE Winter Conference on
  Applications of Computer Vision (WACV)}.\hskip 1em plus 0.5em minus
  0.4em\relax IEEE, 2018, pp. 839--847.

\bibitem{wang2021psspnn}
S.-H. Wang, Y.~Zhang, X.~Cheng, X.~Zhang, and Y.-D. Zhang, ``Psspnn:
  Patchshuffle stochastic pooling neural network for an explainable diagnosis
  of covid-19 with multiple-way data augmentation,'' \emph{Computational and
  Mathematical Methods in Medicine}, vol. 2021, 2021.

\bibitem{song2021deep}
Y.~Song, S.~Zheng, L.~Li, X.~Zhang, X.~Zhang, Z.~Huang, J.~Chen, R.~Wang,
  H.~Zhao, Y.~Zha \emph{et~al.}, ``Deep learning enables accurate diagnosis of
  novel coronavirus (covid-19) with ct images,'' \emph{IEEE/ACM Transactions on
  Computational Biology and Bioinformatics}, 2021.

\bibitem{arias2020artificial}
J.~D. Arias-Londo{\~n}o, J.~A. Gomez-Garcia, L.~Moro-Vel{\'a}zquez, and J.~I.
  Godino-Llorente, ``Artificial intelligence applied to chest x-ray images for
  the automatic detection of covid-19. a thoughtful evaluation approach,''
  \emph{IEEE Access}, vol.~8, pp. 226\,811--226\,827, 2020.

\bibitem{brunese2020explainable}
L.~Brunese, F.~Mercaldo, A.~Reginelli, and A.~Santone, ``Explainable deep
  learning for pulmonary disease and coronavirus covid-19 detection from
  x-rays,'' \emph{Computer Methods and Programs in Biomedicine}, vol. 196, p.
  105608, 2020.

\bibitem{panwar2020deep}
H.~Panwar, P.~Gupta, M.~K. Siddiqui, R.~Morales-Menendez, P.~Bhardwaj, and
  V.~Singh, ``A deep learning and grad-cam based color visualization approach
  for fast detection of covid-19 cases using chest x-ray and ct-scan images,''
  \emph{Chaos, Solitons \& Fractals}, vol. 140, p. 110190, 2020.

\bibitem{alshazly2021explainable}
H.~Alshazly, C.~Linse, E.~Barth, and T.~Martinetz, ``Explainable covid-19
  detection using chest ct scans and deep learning,'' \emph{Sensors}, vol.~21,
  no.~2, p. 455, 2021.

\bibitem{oh2020deep}
Y.~Oh, S.~Park, and J.~C. Ye, ``Deep learning covid-19 features on cxr using
  limited training data sets,'' \emph{IEEE Transactions on Medical Imaging},
  2020.

\bibitem{wang2018non}
X.~Wang, R.~Girshick, A.~Gupta, and K.~He, ``Non-local neural networks,'' in
  \emph{Proceedings of the IEEE conference on computer vision and pattern
  recognition}, 2018, pp. 7794--7803.

\bibitem{chen2019lesion}
B.~Chen, J.~Li, G.~Lu, and D.~Zhang, ``Lesion location attention guided network
  for multi-label thoracic disease classification in chest x-rays,'' \emph{IEEE
  journal of biomedical and health informatics}, 2019.

\bibitem{xu2019pulmonary}
R.~Xu, Z.~Cong, X.~Ye, Y.~Hirano, S.~Kido, T.~Gyobu, Y.~Kawata, O.~Honda, and
  N.~Tomiyama, ``Pulmonary textures classification via a multi-scale attention
  network,'' \emph{IEEE Journal of Biomedical and Health Informatics}, 2019.

\bibitem{chikontwe_dual_2021}
P.~Chikontwe, M.~Luna, M.~Kang, K.~S. Hong, J.~H. Ahn, and S.~H. Park, ``Dual
  attention multiple instance learning with unsupervised complementary loss for
  covid-19 screening,'' \emph{Medical Image Analysis}, p. 102105, 2021.

\bibitem{shi_covid-19_2021}
W.~Shi, L.~Tong, Y.~Zhu, and M.~D. Wang, ``Covid-19 automatic diagnosis with
  radiographic imaging: Explainable attentiontransfer deep neural networks,''
  \emph{IEEE Journal of Biomedical and Health Informatics}, 2021.

\bibitem{dong2021explainable}
H.~Dong, V.~Su{\'a}rez-Paniagua, W.~Whiteley, and H.~Wu, ``Explainable
  automated coding of clinical notes using hierarchical label-wise attention
  networks and label embedding initialisation,'' \emph{Journal of biomedical
  informatics}, vol. 116, p. 103728, 2021.

\bibitem{zhang2018patient2vec}
J.~Zhang, K.~Kowsari, J.~H. Harrison, J.~M. Lobo, and L.~E. Barnes,
  ``Patient2vec: A personalized interpretable deep representation of the
  longitudinal electronic health record,'' \emph{IEEE Access}, vol.~6, pp.
  65\,333--65\,346, 2018.

\bibitem{choi2016retain}
E.~Choi, M.~T. Bahadori, J.~Sun, J.~Kulas, A.~Schuetz, and W.~Stewart,
  ``Retain: An interpretable predictive model for healthcare using reverse time
  attention mechanism,'' \emph{Advances in neural information processing
  systems}, vol.~29, pp. 3504--3512, 2016.

\bibitem{kwon2018retainvis}
B.~C. Kwon, M.-J. Choi, J.~T. Kim, E.~Choi, Y.~B. Kim, S.~Kwon, J.~Sun, and
  J.~Choo, ``Retainvis: Visual analytics with interpretable and interactive
  recurrent neural networks on electronic medical records,'' \emph{IEEE
  transactions on visualization and computer graphics}, vol.~25, no.~1, pp.
  299--309, 2018.

\bibitem{kaji2019attention}
D.~A. Kaji, J.~R. Zech, J.~S. Kim, S.~K. Cho, N.~S. Dangayach, A.~B. Costa, and
  E.~K. Oermann, ``An attention based deep learning model of clinical events in
  the intensive care unit,'' \emph{PloS one}, vol.~14, no.~2, p. e0211057,
  2019.

\bibitem{shickel2019deepsofa}
B.~Shickel, T.~J. Loftus, L.~Adhikari, T.~Ozrazgat-Baslanti, A.~Bihorac, and
  P.~Rashidi, ``Deepsofa: a continuous acuity score for critically ill patients
  using clinically interpretable deep learning,'' \emph{Scientific reports},
  vol.~9, no.~1, pp. 1--12, 2019.

\bibitem{yin2019domain}
C.~Yin, R.~Zhao, B.~Qian, X.~Lv, and P.~Zhang, ``Domain knowledge guided deep
  learning with electronic health records,'' in \emph{2019 IEEE International
  Conference on Data Mining (ICDM)}.\hskip 1em plus 0.5em minus 0.4em\relax
  IEEE, 2019, pp. 738--747.

\bibitem{li2020marrying}
R.~Li, C.~Yin, S.~Yang, B.~Qian, and P.~Zhang, ``Marrying medical domain
  knowledge with deep learning on electronic health records: A deep visual
  analytics approach,'' \emph{Journal of Medical Internet Research}, vol.~22,
  no.~9, p. e20645, 2020.

\bibitem{chen2020interpretable}
P.~Chen, W.~Dong, J.~Wang, X.~Lu, U.~Kaymak, and Z.~Huang, ``Interpretable
  clinical prediction via attention-based neural network,'' \emph{BMC Medical
  Informatics and Decision Making}, vol.~20, no.~3, pp. 1--9, 2020.

\bibitem{lauritsen2020explainable}
S.~M. Lauritsen, M.~Kristensen, M.~V. Olsen, M.~S. Larsen, K.~M. Lauritsen,
  M.~J. J{\o}rgensen, J.~Lange, and B.~Thiesson, ``Explainable artificial
  intelligence model to predict acute critical illness from electronic health
  records,'' \emph{Nature communications}, vol.~11, no.~1, pp. 1--11, 2020.

\bibitem{hegde2019similar}
N.~Hegde, J.~D. Hipp, Y.~Liu, M.~Emmert-Buck, E.~Reif, D.~Smilkov, M.~Terry,
  C.~J. Cai, M.~B. Amin, C.~H. Mermel \emph{et~al.}, ``Similar image search for
  histopathology: Smily,'' \emph{NPJ digital medicine}, vol.~2, no.~1, pp.
  1--9, 2019.

\bibitem{shi2020review}
F.~Shi, J.~Wang, J.~Shi, Z.~Wu, Q.~Wang, Z.~Tang, K.~He, Y.~Shi, and D.~Shen,
  ``Review of artificial intelligence techniques in imaging data acquisition,
  segmentation and diagnosis for covid-19,'' \emph{IEEE reviews in biomedical
  engineering}, 2020.

\bibitem{song2013auto}
C.~Song, F.~Liu, Y.~Huang, L.~Wang, and T.~Tan, ``Auto-encoder based data
  clustering,'' in \emph{Iberoamerican congress on pattern recognition}.\hskip
  1em plus 0.5em minus 0.4em\relax Springer, 2013, pp. 117--124.

\bibitem{lim2020deep}
K.-L. Lim, X.~Jiang, and C.~Yi, ``Deep clustering with variational
  autoencoder,'' \emph{IEEE Signal Processing Letters}, vol.~27, pp. 231--235,
  2020.

\bibitem{prasad2020variational}
V.~Prasad, D.~Das, and B.~Bhowmick, ``Variational clustering: Leveraging
  variational autoencoders for image clustering,'' in \emph{2020 International
  Joint Conference on Neural Networks (IJCNN)}.\hskip 1em plus 0.5em minus
  0.4em\relax IEEE, 2020, pp. 1--10.

\bibitem{king2020unsupervised}
B.~King, S.~Barve, A.~Ford, and R.~Jha, ``Unsupervised clustering of covid-19
  chest x-ray images with a self-organizing feature map,'' in \emph{2020 IEEE
  63rd International Midwest Symposium on Circuits and Systems (MWSCAS)}.\hskip
  1em plus 0.5em minus 0.4em\relax IEEE, 2020, pp. 395--398.

\bibitem{yadav_lung-gans_2021}
P.~Yadav, N.~Menon, V.~Ravi, and S.~Vishvanathan, ``Lung-gans: Unsupervised
  representation learning for lung disease classification using chest ct and
  x-ray images,'' \emph{IEEE Transactions on Engineering Management}, 2021.

\bibitem{singh_these_2021}
G.~Singh and K.-C. Yow, ``These do not look like those: An interpretable deep
  learning model for image recognition,'' \emph{IEEE Access}, vol.~9, pp.
  41\,482--41\,493, 2021.

\bibitem{saeedizadeh2020covid}
N.~Saeedizadeh, S.~Minaee, R.~Kafieh, S.~Yazdani, and M.~Sonka, ``Covid
  tv-unet: Segmenting covid-19 chest ct images using connectivity imposed
  unet,'' \emph{Computer Methods and Programs in Biomedicine Update}, vol.~1,
  p. 100007, 2021.

\bibitem{pennisiAn2021}
M.~Pennisi, I.~Kavasidis, C.~Spampinato, V.~Schinina, S.~Palazzo, F.~P.
  Salanitri, G.~Bellitto, F.~Rundo, M.~Aldinucci, M.~Cristofaro \emph{et~al.},
  ``An explainable ai system for automated covid-19 assessment and lesion
  categorization from ct-scans,'' \emph{Artificial Intelligence in Medicine},
  p. 102114, 2021.

\bibitem{wang_joint_2021}
X.~Wang, L.~Jiang, L.~Li, M.~Xu, X.~Deng, L.~Dai, X.~Xu, T.~Li, Y.~Guo, Z.~Wang
  \emph{et~al.}, ``Joint learning of 3d lesion segmentation and classification
  for explainable covid-19 diagnosis,'' \emph{IEEE Transactions on Medical
  Imaging}, 2021.

\bibitem{yao_csgbbnet_2021}
X.-J. Yao, Z.-Q. Zhu, S.-H. Wang, and Y.-D. Zhang, ``Csgbbnet: An explainable
  deep learning framework for covid-19 detection,'' \emph{Diagnostics},
  vol.~11, no.~9, p. 1712, 2021.

\bibitem{nedumkunnel_explainable_2021}
I.~M. Nedumkunnel, L.~E. George, K.~S. Sowmya, N.~A. Rosh, and V.~Mayya,
  ``Explainable deep neural models for covid-19 prediction from chest x-rays
  with region of interest visualization,'' in \emph{2021 2nd International
  Conference on Secure Cyber Computing and Communications (ICSCCC)}.\hskip 1em
  plus 0.5em minus 0.4em\relax IEEE, 2021, pp. 96--101.

\bibitem{jadhav_covid-view_2021}
S.~Jadhav, G.~Deng, M.~Zawin, and A.~E. Kaufman, ``Covid-view: Diagnosis of
  covid-19 using chest ct,'' \emph{IEEE Transactions on Visualization and
  Computer Graphics}, 2021.

\bibitem{natekar2020demystifiying}
P.~Natekar, A.~Kori, and G.~Krishnamurthi, ``Demystifying brain tumor
  segmentation networks: interpretability and uncertainty analysis,''
  \emph{Frontiers in computational neuroscience}, vol.~14, p.~6, 2020.

\bibitem{vaid2020machine}
A.~Vaid, S.~Somani, A.~J. Russak, J.~K. De~Freitas, F.~F. Chaudhry,
  I.~Paranjpe, K.~W. Johnson, S.~J. Lee, R.~Miotto, F.~Richter \emph{et~al.},
  ``Machine learning to predict mortality and critical events in a cohort of
  patients with covid-19 in new york city: Model development and validation,''
  \emph{Journal of medical Internet research}, vol.~22, no.~11, p. e24018,
  2020.

\bibitem{wollenstein2020physiological}
S.~Wollenstein-Betech, A.~A. Silva, J.~L. Fleck, C.~G. Cassandras, and I.~C.
  Paschalidis, ``Physiological and socioeconomic characteristics predict
  covid-19 mortality and resource utilization in brazil,'' \emph{PloS one},
  vol.~15, no.~10, p. e0240346, 2020.

\bibitem{pan2020prognostic}
P.~Pan, Y.~Li, Y.~Xiao, B.~Han, L.~Su, M.~Su, Y.~Li, S.~Zhang, D.~Jiang,
  X.~Chen \emph{et~al.}, ``Prognostic assessment of covid-19 in the intensive
  care unit by machine learning methods: Model development and validation,''
  \emph{Journal of medical Internet research}, vol.~22, no.~11, p. e23128,
  2020.

\bibitem{lu2020explainable}
J.~Lu, R.~Jin, E.~Song, M.~Alrashoud, K.~N. Al-Mutib, and M.~S. Al-Rakhami,
  ``An explainable system for diagnosis and prognosis of covid-19,'' \emph{IEEE
  Internet of Things Journal}, 2020.

\bibitem{sanchez2020machine}
M.~S{\'a}nchez-Monta{\~n}{\'e}s, P.~Rodr{\'\i}guez-Belenguer, A.~J.
  Serrano-L{\'o}pez, E.~Soria-Olivas, and Y.~Alakhdar-Mohmara, ``Machine
  learning for mortality analysis in patients with covid-19,''
  \emph{International journal of environmental research and public health},
  vol.~17, no.~22, p. 8386, 2020.

\bibitem{brinati2020detection}
D.~Brinati, A.~Campagner, D.~Ferrari, M.~Locatelli, G.~Banfi, and F.~Cabitza,
  ``Detection of covid-19 infection from routine blood exams with machine
  learning: a feasibility study,'' \emph{Journal of medical systems}, vol.~44,
  no.~8, pp. 1--12, 2020.

\bibitem{alves2021explaining}
M.~A. Alves, G.~Z. Castro, B.~A.~S. Oliveira, L.~A. Ferreira, J.~A.
  Ram{\'\i}rez, R.~Silva, and F.~G. Guimar{\~a}es, ``Explaining machine
  learning based diagnosis of covid-19 from routine blood tests with decision
  trees and criteria graphs,'' \emph{Computers in Biology and Medicine}, vol.
  132, p. 104335, 2021.

\bibitem{estiri2021individualized}
H.~Estiri, Z.~H. Strasser, and S.~N. Murphy, ``Individualized prediction of
  covid-19 adverse outcomes with mlho,'' \emph{Scientific reports}, vol.~11,
  no.~1, pp. 1--9, 2021.

\bibitem{qomariyah2021tree}
N.~N. Qomariyah, A.~A. Purwita, S.~D.~A. Asri, and D.~Kazakov, ``A tree-based
  mortality prediction model of covid-19 from routine blood samples,'' in
  \emph{2021 International Conference on ICT for Smart Society (ICISS)}.\hskip
  1em plus 0.5em minus 0.4em\relax IEEE, 2021, pp. 1--7.

\bibitem{famiglini2021prediction}
L.~Famiglini, G.~Bini, A.~Carobene, A.~Campagner, and F.~Cabitza, ``Prediction
  of icu admission for covid-19 patients: a machine learning approach based on
  complete blood count data,'' in \emph{2021 IEEE 34th International Symposium
  on Computer-Based Medical Systems (CBMS)}.\hskip 1em plus 0.5em minus
  0.4em\relax IEEE, 2021, pp. 160--165.

\bibitem{ou2020rupture}
C.~Ou, J.~Liu, Y.~Qian, W.~Chong, X.~Zhang, W.~Liu, H.~Su, N.~Zhang, J.~Zhang,
  C.-Z. Duan \emph{et~al.}, ``Rupture risk assessment for cerebral aneurysm
  using interpretable machine learning on multidimensional data,''
  \emph{Frontiers in neurology}, vol.~11, p. 1696, 2020.

\bibitem{liu2018interpretable}
L.~Liu, Y.~Yu, Z.~Fei, M.~Li, F.-X. Wu, H.-D. Li, Y.~Pan, and J.~Wang, ``An
  interpretable boosting model to predict side effects of analgesics for
  osteoarthritis,'' \emph{BMC systems biology}, vol.~12, no.~6, pp. 29--38,
  2018.

\bibitem{song2020cross}
X.~Song, S.~Alan, J.~A. Kellum, L.~R. Waitman, M.~E. Matheny, S.~Q. Simpson,
  Y.~Hu, and M.~Liu, ``Cross-site transportability of an explainable artificial
  intelligence model for acute kidney injury prediction,'' \emph{Nature
  communications}, vol.~11, no.~1, pp. 1--12, 2020.

\bibitem{yang2016predicting}
C.~Yang, C.~Delcher, E.~Shenkman, and S.~Ranka, ``Predicting 30-day all-cause
  readmissions from hospital inpatient discharge data,'' in \emph{2016 IEEE
  18th International conference on e-Health networking, applications and
  services (Healthcom)}.\hskip 1em plus 0.5em minus 0.4em\relax IEEE, 2016, pp.
  1--6.

\bibitem{misra2021early}
D.~Misra, V.~Avula, D.~M. Wolk, H.~A. Farag, J.~Li, Y.~B. Mehta, R.~Sandhu,
  B.~Karunakaran, S.~Kethireddy, R.~Zand \emph{et~al.}, ``Early detection of
  septic shock onset using interpretable machine learners,'' \emph{Journal of
  Clinical Medicine}, vol.~10, no.~2, p. 301, 2021.

\bibitem{che2016interpretable}
Z.~Che, S.~Purushotham, R.~Khemani, and Y.~Liu, ``Interpretable deep models for
  icu outcome prediction,'' in \emph{AMIA annual symposium proceedings}, vol.
  2016.\hskip 1em plus 0.5em minus 0.4em\relax American Medical Informatics
  Association, 2016, p. 371.

\bibitem{dupuis2020limited}
C.~Dupuis, E.~De~Montmollin, M.~Neuville, B.~Mourvillier, S.~Ruckly, and
  J.~Timsit, ``Limited applicability of a covid-19 specific mortality
  prediction rule to the intensive care setting,'' \emph{Nature Machine
  Intelligence}, pp. 1--3, 2020.

\bibitem{quanjel2020replication}
M.~J. Quanjel, T.~C. van Holten, P.~C. Gunst-van~der Vliet, J.~Wielaard,
  B.~Karakaya, M.~S{\"o}hne, H.~S. Moeniralam, and J.~C. Grutters,
  ``Replication of a mortality prediction model in dutch patients with
  covid-19,'' \emph{Nature Machine Intelligence}, pp. 1--2, 2020.

\bibitem{barish2020external}
M.~Barish, S.~Bolourani, L.~F. Lau, S.~Shah, and T.~P. Zanos, ``External
  validation demonstrates limited clinical utility of the interpretable
  mortality prediction model for patients with covid-19,'' \emph{Nature Machine
  Intelligence}, pp. 1--3, 2020.

\bibitem{torres2020pandemyc}
J.~Torres-Macho, P.~Ryan, J.~Valencia, M.~P{\'e}rez-Butrague{\~n}o,
  E.~Jim{\'e}nez, M.~Font{\'a}n-Vela, E.~Izquierdo-Garc{\'\i}a,
  I.~Fernandez-Jimenez, E.~{\'A}lvaro-Alonso, A.~Lazaro \emph{et~al.}, ``The
  pandemyc score. an easily applicable and interpretable model for predicting
  mortality associated with covid-19,'' \emph{Journal of clinical medicine},
  vol.~9, no.~10, p. 3066, 2020.

\bibitem{foieni2020derivation}
F.~Foieni, G.~Sala, J.~G. Mognarelli, G.~Suigo, D.~Zampini, M.~Pistoia,
  M.~Ciola, T.~Ciampani, C.~Ultori, and P.~Ghiringhelli, ``Derivation and
  validation of the clinical prediction model for covid-19,'' \emph{Internal
  and Emergency Medicine}, vol.~15, no.~8, pp. 1409--1414, 2020.

\bibitem{tang2020prediction}
G.~Tang, Y.~Luo, F.~Lu, W.~Li, X.~Liu, Y.~Nan, Y.~Ren, X.~Liao, S.~Wu, H.~Jin
  \emph{et~al.}, ``Prediction of sepsis in covid-19 using laboratory
  indicators,'' \emph{Frontiers in cellular and infection microbiology},
  vol.~10, 2020.

\bibitem{zheng2021interpretable}
B.~Zheng, Y.~Cai, F.~Zeng, M.~Lin, J.~Zheng, W.~Chen, G.~Qin, and Y.~Guo, ``An
  interpretable model-based prediction of severity and crucial factors in
  patients with covid-19,'' \emph{BioMed Research International}, vol. 2021,
  2021.

\bibitem{makridis2021designing}
C.~A. Makridis, T.~Strebel, V.~Marconi, and G.~Alterovitz, ``Designing covid-19
  mortality predictions to advance clinical outcomes: Evidence from the
  department of veterans affairs,'' \emph{BMJ Health \& Care Informatics},
  vol.~28, no.~1, 2021.

\bibitem{haimovich2020development}
A.~D. Haimovich, N.~G. Ravindra, S.~Stoytchev, H.~P. Young, F.~P. Wilson,
  D.~van Dijk, W.~L. Schulz, and R.~A. Taylor, ``Development and validation of
  the quick covid-19 severity index: a prognostic tool for early clinical
  decompensation,'' \emph{Annals of emergency medicine}, vol.~76, no.~4, pp.
  442--453, 2020.

\bibitem{abdulaal2020prognostic}
A.~Abdulaal, A.~Patel, E.~Charani, S.~Denny, N.~Mughal, L.~Moore \emph{et~al.},
  ``Prognostic modeling of covid-19 using artificial intelligence in the united
  kingdom: model development and validation,'' \emph{Journal of Medical
  Internet Research}, vol.~22, no.~8, p. e20259, 2020.

\bibitem{cavallaro2021contrasting}
M.~Cavallaro, H.~Moiz, M.~J. Keeling, and N.~D. McCarthy, ``Contrasting factors
  associated with covid-19-related icu admission and death outcomes in
  hospitalised patients by means of shapley values,'' \emph{medRxiv}, pp.
  2020--12, 2021.

\bibitem{patel2021natural}
B.~V. Patel, S.~Haar, R.~Handslip, C.~Auepanwiriyakul, T.~M.-L. Lee, S.~Patel,
  J.~A. Harston, F.~Hosking-Jervis, D.~Kelly, B.~Sanderson \emph{et~al.},
  ``Natural history, trajectory, and management of mechanically ventilated
  covid-19 patients in the united kingdom,'' \emph{Intensive care medicine},
  vol.~47, no.~5, pp. 549--565, 2021.

\bibitem{magunia2021machine}
H.~Magunia, S.~Lederer, R.~Verbuecheln, B.~J. Gilot, M.~Koeppen, H.~A.
  Haeberle, V.~Mirakaj, P.~Hofmann, G.~Marx, J.~Bickenbach \emph{et~al.},
  ``Machine learning identifies icu outcome predictors in a multicenter
  covid-19 cohort,'' \emph{Critical Care}, vol.~25, no.~1, pp. 1--14, 2021.

\bibitem{singhal2021eards}
L.~Singhal, Y.~Garg, P.~Yang, A.~Tabaie, A.~I. Wong, A.~Mohammed, L.~Chinthala,
  D.~Kadaria, A.~Sodhi, A.~L. Holder \emph{et~al.}, ``eards: A multi-center
  validation of an interpretable machine learning algorithm of early onset
  acute respiratory distress syndrome (ards) among critically ill adults with
  covid-19,'' \emph{PloS one}, vol.~16, no.~9, p. e0257056, 2021.

\bibitem{nguyen2021budget}
S.~Nguyen, R.~Chan, J.~Cadena, B.~Soper, P.~Kiszka, L.~Womack, M.~Work, J.~M.
  Duggan, S.~T. Haller, J.~A. Hanrahan \emph{et~al.}, ``Budget constrained
  machine learning for early prediction of adverse outcomes for covid-19
  patients,'' \emph{Scientific Reports}, vol.~11, no.~1, pp. 1--14, 2021.

\bibitem{rodriguez2021development}
V.~A. Rodriguez, S.~Bhave, R.~Chen, C.~Pang, G.~Hripcsak, S.~Sengupta,
  N.~Elhadad, R.~Green, J.~Adelman, K.~S. Metitiri \emph{et~al.}, ``Development
  and validation of prediction models for mechanical ventilation, renal
  replacement therapy, and readmission in covid-19 patients,'' \emph{Journal of
  the American Medical Informatics Association: JAMIA}, 2021.

\bibitem{wu2021interpretable}
H.~Wu, W.~Ruan, J.~Wang, D.~Zheng, B.~Liu, Y.~Geng, X.~Chai, J.~Chen, K.~Li,
  S.~Li \emph{et~al.}, ``Interpretable machine learning for covid-19: an
  empirical study on severity prediction task,'' \emph{IEEE Transactions on
  Artificial Intelligence}, 2021.

\bibitem{martini2021drivers}
M.~L. Martini, S.~N. Neifert, J.~S. Gal, E.~K. Oermann, J.~T. Gilligan, and
  J.~M. Caridi, ``Drivers of prolonged hospitalization following spine surgery:
  A game-theory-based approach to explaining machine learning models,''
  \emph{JBJS}, vol. 103, no.~1, pp. 64--73, 2021.

\bibitem{rodriguez2019interpretation}
R.~Rodr{\'\i}guez-P{\'e}rez and J.~Bajorath, ``Interpretation of compound
  activity predictions from complex machine learning models using local
  approximations and shapley values,'' \emph{Journal of Medicinal Chemistry},
  vol.~63, no.~16, pp. 8761--8777, 2019.

\bibitem{dai2020development}
C.~Dai, Y.~Fan, Y.~Li, X.~Bao, Y.~Li, M.~Su, Y.~Yao, K.~Deng, B.~Xing, F.~Feng
  \emph{et~al.}, ``Development and interpretation of multiple machine learning
  models for predicting postoperative delayed remission of acromegaly patients
  during long-term follow-up,'' \emph{Frontiers in endocrinology}, vol.~11,
  2020.

\bibitem{caicedo2019iseeu}
W.~Caicedo-Torres and J.~Gutierrez, ``Iseeu: Visually interpretable deep
  learning for mortality prediction inside the icu,'' \emph{Journal of
  biomedical informatics}, vol.~98, p. 103269, 2019.

\bibitem{fisher2019all}
A.~Fisher, C.~Rudin, and F.~Dominici, ``All models are wrong, but many are
  useful: Learning a variable's importance by studying an entire class of
  prediction models simultaneously.'' \emph{J. Mach. Learn. Res.}, vol.~20, no.
  177, pp. 1--81, 2019.

\bibitem{cao2020impact}
Z.~Cao, F.~Tang, C.~Chen, C.~Zhang, Y.~Guo, R.~Lin, Z.~Huang, Y.~Teng, T.~Xie,
  Y.~Xu \emph{et~al.}, ``Impact of systematic factors on the outbreak outcomes
  of the novel covid-19 disease in china: factor analysis study,''
  \emph{Journal of medical Internet research}, vol.~22, no.~11, p. e23853,
  2020.

\bibitem{ramchandani2020deepcovidnet}
A.~Ramchandani, C.~Fan, and A.~Mostafavi, ``Deepcovidnet: An interpretable deep
  learning model for predictive surveillance of covid-19 using heterogeneous
  features and their interactions,'' \emph{IEEE Access}, vol.~8, pp.
  159\,915--159\,930, 2020.

\bibitem{patelNatural2021}
B.~V. Patel, S.~Haar, R.~Handslip, C.~Auepanwiriyakul, T.~M.-L. Lee, S.~Patel,
  J.~A. Harston, F.~Hosking-Jervis, D.~Kelly, B.~Sanderson \emph{et~al.},
  ``Natural history, trajectory, and management of mechanically ventilated
  covid-19 patients in the united kingdom,'' \emph{Intensive care medicine},
  vol.~47, no.~5, pp. 549--565, 2021.

\bibitem{tsiknakis2020interpretable}
N.~Tsiknakis, E.~Trivizakis, E.~E. Vassalou, G.~Z. Papadakis, D.~A. Spandidos,
  A.~Tsatsakis, J.~S{\'a}nchez-Garc{\'\i}a, R.~L{\'o}pez-Gonz{\'a}lez,
  N.~Papanikolaou, A.~H. Karantanas \emph{et~al.}, ``Interpretable artificial
  intelligence framework for covid-19 screening on chest x-rays,''
  \emph{Experimental and Therapeutic Medicine}, vol.~20, no.~2, pp. 727--735,
  2020.

\bibitem{samek2016evaluating}
W.~Samek, A.~Binder, G.~Montavon, S.~Lapuschkin, and K.-R. M{\"u}ller,
  ``Evaluating the visualization of what a deep neural network has learned,''
  \emph{IEEE transactions on neural networks and learning systems}, vol.~28,
  no.~11, pp. 2660--2673, 2016.

\bibitem{petsiuk2018rise}
V.~Petsiuk, A.~Das, and K.~Saenko, ``Rise: Randomized input sampling for
  explanation of black-box models,'' \emph{arXiv preprint arXiv:1806.07421},
  2018.

\bibitem{devries2019evaluation}
T.~DeVries, A.~Romero, L.~Pineda, G.~W. Taylor, and M.~Drozdzal, ``On the
  evaluation of conditional gans,'' \emph{arXiv preprint arXiv:1907.08175},
  2019.

\bibitem{yang2019bim}
M.~Yang and B.~Kim, ``Bim: Towards quantitative evaluation of interpretability
  methods with ground truth,'' \emph{arXiv preprint arXiv:1907.09701}, 2019.

\bibitem{10.1145/3447548.3467213}
Y.-S. Lin, W.-C. Lee, and Z.~B. Celik, ``What do you see? evaluation of
  explainable artificial intelligence (xai) interpretability through neural
  backdoors,'' in \emph{Proceedings of the 27th ACM SIGKDD Conference on
  Knowledge Discovery \& Data Mining}, ser. KDD '21, 2021, p. 1027–1035.

\bibitem{schirrmeister2017deep}
R.~T. Schirrmeister, J.~T. Springenberg, L.~D.~J. Fiederer, M.~Glasstetter,
  K.~Eggensperger, M.~Tangermann, F.~Hutter, W.~Burgard, and T.~Ball, ``Deep
  learning with convolutional neural networks for eeg decoding and
  visualization,'' \emph{Human brain mapping}, vol.~38, no.~11, pp. 5391--5420,
  2017.

\bibitem{hossain2019applying}
M.~S. Hossain, S.~U. Amin, M.~Alsulaiman, and G.~Muhammad, ``Applying deep
  learning for epilepsy seizure detection and brain mapping visualization,''
  \emph{ACM Transactions on Multimedia Computing, Communications, and
  Applications (TOMM)}, vol.~15, no.~1s, pp. 1--17, 2019.

\bibitem{vaswani2017attention}
A.~Vaswani, N.~Shazeer, N.~Parmar, J.~Uszkoreit, L.~Jones, A.~N. Gomez,
  {\L}.~Kaiser, and I.~Polosukhin, ``Attention is all you need,'' in
  \emph{Advances in neural information processing systems}, 2017, pp.
  5998--6008.

\bibitem{pennisi2021explainable}
M.~Pennisi, I.~Kavasidis, C.~Spampinato, V.~Schinina, S.~Palazzo, F.~P.
  Salanitri, G.~Bellitto, F.~Rundo, M.~Aldinucci, M.~Cristofaro \emph{et~al.},
  ``An explainable ai system for automated covid-19 assessment and lesion
  categorization from ct-scans,'' \emph{Artificial Intelligence in Medicine},
  p. 102114, 2021.

\bibitem{brinatiDetection2020}
D.~Brinati, A.~Campagner, D.~Ferrari, M.~Locatelli, G.~Banfi, and F.~Cabitza,
  ``Detection of covid-19 infection from routine blood exams with machine
  learning: a feasibility study,'' \emph{Journal of medical systems}, vol.~44,
  no.~8, pp. 1--12, 2020.

\bibitem{makridisDesigning2021}
C.~A. Makridis, T.~Strebel, V.~Marconi, and G.~Alterovitz, ``Designing covid-19
  mortality predictions to advance clinical outcomes: Evidence from the
  department of veterans affairs,'' \emph{BMJ Health \& Care Informatics},
  vol.~28, no.~1, 2021.

\bibitem{hou_explainable_2021}
J.~Hou and T.~Gao, ``Explainable dcnn based chest x-ray image analysis and
  classification for covid-19 pneumonia detection,'' \emph{Scientific Reports},
  vol.~11, no.~1, pp. 1--15, 2021.

\bibitem{Zheng2019-bt}
Q.~Zheng, H.~Delingette, and N.~Ayache, ``Explainable cardiac pathology
  classification on cine mri with motion characterization by semi-supervised
  learning of apparent flow,'' \emph{Medical image analysis}, vol.~56, pp.
  80--95, 2019.

\bibitem{jones2009sequential}
A.~E. Jones, S.~Trzeciak, and J.~A. Kline, ``The sequential organ failure
  assessment score for predicting outcome in patients with severe sepsis and
  evidence of hypoperfusion at the time of emergency department presentation,''
  \emph{Critical care medicine}, vol.~37, no.~5, p. 1649, 2009.

\bibitem{barlow2007curb65}
G.~Barlow, D.~Nathwani, and P.~Davey, ``The curb65 pneumonia severity score
  outperforms generic sepsis and early warning scores in predicting mortality
  in community-acquired pneumonia,'' \emph{Thorax}, vol.~62, no.~3, pp.
  253--259, 2007.

\bibitem{chicco2020advantages}
D.~Chicco and G.~Jurman, ``The advantages of the matthews correlation
  coefficient (mcc) over f1 score and accuracy in binary classification
  evaluation,'' \emph{BMC genomics}, vol.~21, no.~1, pp. 1--13, 2020.

\bibitem{argaw2020cybersecurity}
S.~T. Argaw, J.~R. Troncoso-Pastoriza, D.~Lacey, M.-V. Florin, F.~Calcavecchia,
  D.~Anderson, W.~Burleson, J.-M. Vogel, C.~O’Leary, B.~Eshaya-Chauvin
  \emph{et~al.}, ``Cybersecurity of hospitals: discussing the challenges and
  working towards mitigating the risks,'' \emph{BMC Medical Informatics and
  Decision Making}, vol.~20, no.~1, pp. 1--10, 2020.

\bibitem{kang_quantitative_2021}
M.~Kang, K.~S. Hong, P.~Chikontwe, M.~Luna, J.~G. Jang, J.~Park, K.-C. Shin,
  S.~H. Park, and J.~H. Ahn, ``Quantitative assessment of chest ct patterns in
  covid-19 and bacterial pneumonia patients: a deep learning perspective,''
  \emph{Journal of Korean medical science}, vol.~36, no.~5, 2021.

\bibitem{Bau2020-rx}
D.~Bau, J.-Y. Zhu, H.~Strobelt, A.~Lapedriza, B.~Zhou, and A.~Torralba,
  ``Understanding the role of individual units in a deep neural network,''
  \emph{Proceedings of the National Academy of Sciences}, vol. 117, no.~48, pp.
  30\,071--30\,078, 2020.

\end{thebibliography}

\end{document}